\documentclass[11pt,a4paper]{article}

\usepackage[margin=1in]{geometry}
\usepackage{amsmath,amssymb,amsfonts}
\usepackage{mathtools}
\usepackage{booktabs}
\usepackage{array}
\usepackage{multirow}
\usepackage{tabularx}
\usepackage{graphicx}
\usepackage{float}
\usepackage{tcolorbox}
\usepackage{booktabs}
\tcbuselibrary{skins, breakable}
\usepackage{graphicx}
\usepackage{float} 

\usepackage{xcolor}
\usepackage{hyperref}
\usepackage{url}
\usepackage{microtype}
\usepackage{enumitem}
\usepackage{caption}
\usepackage{subcaption}
\usepackage{tikz}
\usepackage{pgfplots}
\pgfplotsset{compat=1.18}
\usetikzlibrary{shapes.geometric, arrows.meta, positioning, fit, backgrounds, calc}
\usepackage{algorithm}
\usepackage{algpseudocode}
\usepackage{natbib}
\usepackage{setspace}
\usepackage{titlesec}
\usepackage{fancyhdr}
\usepackage{listings}
\usepackage{tcolorbox}
\tcbuselibrary{breakable, skins}

\newtcolorbox{promptbox}[1][]{%
  breakable,
  enhanced,
  colback=lightgray,
  colframe=black,
  fonttitle=\bfseries\small,
  title={#1},
  left=6pt, right=6pt, top=4pt, bottom=4pt,
  boxrule=0.6pt,
  arc=3pt
}
\usepackage{mdframed}

\hypersetup{
  colorlinks=true,
  linkcolor=black,
  citecolor=black,
  urlcolor=blue,
  filecolor=black,
  pdftitle={PassiveQA: A Three-Action Framework for Epistemically Calibrated Question Answering via Supervised Finetuning},
  pdfauthor={Madhav S Baidya}
}

\definecolor{varnodecolor}{RGB}{210,228,248}
\definecolor{lightgray}{RGB}{245,245,245}
\definecolor{darkgray}{RGB}{80,80,80}
\definecolor{varnode}{RGB}{220,230,245}   
\definecolor{ansedge}{RGB}{180,220,180}   
\definecolor{abstainedge}{RGB}{230,200,200} 

\titleformat{\section}{\large\bfseries}{\thesection}{1em}{}
\titleformat{\subsection}{\normalsize\bfseries}{\thesubsection}{1em}{}
\titleformat{\subsubsection}{\normalsize\itshape}{\thesubsubsection}{1em}{}

\tikzset{
  block/.style={rectangle, rounded corners=2pt, draw=black, thick,
                fill=lightgray, text width=2.8cm, align=center,
                minimum height=0.8cm, font=\small},
  decision/.style={diamond, draw=black, thick, fill=lightgray,
                   aspect=2.5, align=center, font=\footnotesize},
  arrow/.style={-{Stealth[length=5pt]}, thick},
  dasharrow/.style={-{Stealth[length=5pt]}, thick, dashed},
  entity/.style={rectangle, rounded corners=3pt, draw=black, thick,
                 fill=lightgray, align=center, font=\footnotesize,
                 minimum height=0.65cm, inner sep=4pt},
  varnode/.style={rectangle, rounded corners=3pt, draw=black!60, thick,
                  fill=varnode, align=center, font=\footnotesize\itshape,
                  minimum height=0.65cm, inner sep=4pt},
  kgphase/.style={rectangle, draw=black!40, rounded corners=4pt,
                  inner sep=8pt, fill=white},
  kedge/.style={-{Stealth[length=4pt]}, thick},
  kedge_ans/.style={-{Stealth[length=4pt]}, thick, draw=black!70,
                    line width=1.5pt},
  kedge_abs/.style={-{Stealth[length=4pt]}, thick, draw=black!30, dashed},
  kedge_var/.style={-{Stealth[length=4pt]}, thick, draw=black!60,
                    line width=1pt, dotted}
}

\begin{document}

\title{%
  \textbf{PassiveQA: A Three-Action Framework for Epistemically Calibrated\\
  Question Answering via Supervised Finetuning}
}

\author{%
  Madhav S Baidya\\[4pt]
  \small Indian Institute of Technology (BHU) Varanasi\\
  \small \texttt{madhavsukla.baidya.chy22@itbhu.ac.in}
}

\date{}
\maketitle
\thispagestyle{fancy}

\begin{abstract}
Large Language Models (LLMs) have achieved strong performance in question answering
and retrieval-augmented generation (RAG), yet they implicitly assume that user queries are
fully specified and answerable. In real-world settings, queries are often incomplete,
ambiguous, or missing critical variables, leading models to produce overconfident or
hallucinated responses. Existing approaches lack mechanisms to decide whether a query
should be answered, clarified, or abstained from.

In this work, we study \emph{decision-aware query resolution under incomplete
information}, where a model must jointly infer the sufficiency of available information
and select an appropriate action: \textsc{Answer}, \textsc{Ask}, or \textsc{Abstain}. We
show that standard and enhanced RAG systems do not consistently exhibit such epistemic awareness,
defaulting to answer generation even when information is insufficient.

To address this, we propose \textbf{PassiveQA}, a three-action decision framework that
learns to align model behaviour with information sufficiency through supervised finetuning.
Our approach integrates structured state representations, knowledge graph-grounded
context, and a finetuned planner that explicitly models missing variables and decision
reasoning.

Experiments across four multi-source QA datasets show that the finetuned 
planner achieves 55.6\% macro F1 — a 20.3 percentage point gain over the 
best inference-time architecture — with \textsc{Abstain} recall rising 
from 13.3\% to 58.1\% and hallucination rate falling from 42.7\% to 
33.8\%, while maintaining 100\% structured output compliance across 
5{,}218 test samples.
Our results highlight the limitations of inference-time interventions and emphasise the need
for training-time alignment of epistemic decision-making in language models.
\end{abstract}


\vspace{0.6em}
\noindent\textbf{Code Availability.}\;
The full codebase — including dataset construction pipelines, knowledge graph
builder, finetuning dataset generator, LoRA training scripts, and the
three-agent inference pipeline — is publicly available at
\url{https://github.com/MadsDoodle/PassiveQA}.
Questions, bug reports, and suggestions are welcome via the repository's
issue tracker.

\vspace{0.5em}
\noindent\rule{\textwidth}{0.4pt}

\section{Introduction}
\label{sec:intro}

Consider a legal QA system asked: \emph{``Am I entitled to redundancy pay?''}
The system retrieves the correct statutory clause and responds: \emph{``Yes,
employees are entitled to statutory redundancy pay after two years of continuous
service''} — without ever establishing that the user has been employed for
fourteen months.  The retrieved document is correct.  The generated text is
faithful to it.  The system fails anyway, because it committed to an answer
without first determining whether the information available was sufficient to
ground one.  This is an \emph{epistemic failure}, structurally invisible to
standard retrieval and generation metrics.

Large language models and retrieval-augmented generation systems share an
assumption that is rarely stated and almost never challenged: every query is
answerable~\citep{lewis2020rag, gao2023rag_survey}.  The system's only task,
in the standard framing, is to find the best answer.  There is no mechanism
for deciding whether an answer should be attempted at all.  This assumption
fails routinely in deployment — queries arrive incomplete, with implicit
variables the user expects the system to ask about, or with topics absent from
the knowledge base entirely.  In each case, the standard RAG response is the
same: retrieve the highest-scoring chunks and generate the most fluent
completion~\citep{hallucination_mitigating_survey, ji2023hallucination}.
Critically, improving retrieval quality does not fix this.  Better retrieval
surfaces more plausible-looking evidence, which, without an epistemic decision
gate, \emph{increases} overconfident
answering~\citep{hallucination_mitigating_survey}.  We demonstrate this
empirically: an enhanced RAG pipeline with semantic chunking, hybrid retrieval,
cross-encoder reranking, and self-reflection achieves a hallucination rate of
51.7\% — higher than the 42.7\% baseline it was designed to improve.

The standard mitigation is confidence
calibration~\citep{geifman2017selective, xin2021abstention} — assign a scalar
to each response and abstain below a threshold.  This framing is insufficient
for two reasons.  First, it collapses two qualitatively distinct epistemic
states: a query unanswerable because its topic is absent from the KB
(irrecoverable — honest refusal is correct) versus a query unanswerable
because a critical variable is missing but \emph{recoverable through dialogue}
(a targeted clarification question is correct).  Conflating these discards
information the user needs~\citep{aliannejadi2019clarify, saeidi2018sharc}.
Second, entropy-based confidence signals are insufficient for safe selective
prediction in LLMs~\citep{phillips2026entropy}, and models frequently fail to
refuse even when their confidence is
low~\citep{yin2024llm_abstention}.

We instead ground the routing decision in the \emph{information state}
$S(q) = (V_{\mathrm{known}},\, V_{\mathrm{missing}},\, C)$ and define
three actions with non-overlapping epistemic semantics:
\begin{enumerate}[leftmargin=*, label=(\roman*), itemsep=2pt]
  \item \textbf{\textsc{Answer}} — $V_{\mathrm{missing}} = \emptyset$ and
        retrieved evidence provides a complete reasoning path.
  \item \textbf{\textsc{Ask}} — $V_{\mathrm{missing}} \neq \emptyset$ but
        missing variables are \emph{recoverable}: a targeted clarification
        resolves the gap.
  \item \textbf{\textsc{Abstain}} — $V_{\mathrm{missing}} \neq \emptyset$
        and missing information is \emph{irrecoverable}: no user clarification
        will help.
\end{enumerate}
Hallucination risk scales monotonically with the incompleteness measure
$I(q) = |V_{\mathrm{missing}}| / (|V_{\mathrm{known}}| +
|V_{\mathrm{missing}}|)$, motivating variable-state tracking as the primary
routing signal.

The central claim of this paper is that this epistemic calibration
\emph{cannot be achieved at inference time}.  We establish this by exhausting
the space of inference-time interventions across three progressively stronger
RAG architectures.  All three plateau at 34--38\% decision accuracy and below
14\% \textsc{Abstain} recall, with a consistent failure mode: the model
defaults to \textsc{Answer} regardless of the information state, because that
behaviour is encoded in its pretraining
distribution~\citep{ouyang2022rlhf}.  No inference-time signal can override a
prior this strong without modifying the parameters that encode it.

To address this, we propose \textbf{PassiveQA}, a framework for
epistemically calibrated QA through training-time alignment.  PassiveQA
constructs a \emph{decision-weighted knowledge graph} $G_2$ in which edge
weights encode the three-action supervision signal — \textsc{Answer}-supporting
paths are reinforced, \textsc{Abstain}-associated paths penalised, and
recoverable missing variables are injected as explicit \texttt{?var}
placeholder nodes.  A 34K-sample finetuning corpus of graph-grounded
structured reasoning chains then trains a LoRA-adapted Mistral-7B-Instruct
planner~\citep{jiang2023mistral, hu2021lora} to produce explicit, structured
decisions over the three-action space.

\paragraph{Results.}
The finetuned planner achieves \textbf{55.6\% macro F1} on the held-out test
split — a \textbf{20.3~pp gain} over the best inference-time architecture
(v3: 35.3\%) and \textbf{28.9~pp} over the standard RAG baseline (26.7\%).
\textsc{Abstain} recall rises from 13.3\% to \textbf{58.1\%}, confirming that
the empty-graph signal is learnable through training in a way that
inference-time thresholds cannot replicate.  Per-output structural compliance
is 100\% across 5{,}218 test samples.  These results hold under a severely
compute-constrained regime — 2 epochs over 9{,}000 samples (26\% of the
available 34K finetuning set) — establishing a conservative lower bound.

\paragraph{Positioning.}
PassiveQA sits at the intersection of three research threads that have
largely developed independently: selective prediction~\citep{geifman2017selective,
xin2021abstention}, conversational clarification~\citep{saeidi2018sharc,
aliannejadi2019clarify}, and knowledge graph-augmented
reasoning~\citep{yasunaga2021qagnn, sun2019kgqa}.  Prior work in selective
prediction operates on classification models with calibrated outputs and
treats abstention as a binary post-hoc decision.  Prior work on clarification
focuses on question generation conditioned on a fixed dialogue policy, without
modelling the information state that determines \emph{whether} to ask.  Prior
work on KG-augmented QA uses graph structure to improve answer quality, but
not to route queries away from answering entirely.  PassiveQA unifies these
threads by treating the three-action routing decision as a first-class
learning objective, grounded simultaneously in structured knowledge, explicit
variable-state tracking, and training-time behavioural alignment.  The
decision-weighted knowledge graph $G_2$ is the concrete artefact that makes
this unification possible: it is the only component in the pipeline that
jointly encodes semantic grounding, multi-hop reasoning signal, and
epistemic supervision in a single structure that the planner can query at
inference time.

\noindent\textbf{Contributions:}
\begin{itemize}[leftmargin=*, itemsep=2pt]
  \item A formal three-action decision framework grounded in decision theory
        and information-state decomposition, including a joint answerability
        signal $A(q)$ and multi-turn resolution rate $\rho_t$
        (\S\ref{sec:formalism}).
  \item Strong empirical evidence that inference-time interventions fail to consistently instil epistemic passivity, established across three progressively stronger
        RAG architectures (\S\ref{sec:rag}).
  \item A multi-source dataset pipeline unifying four QA benchmarks with
        explicit variable-state supervision (\S\ref{sec:data}).
  \item A three-phase KG construction procedure whose central novelty is
        query-guided decision reinforcement on edge weights and \texttt{?var}
        placeholder injection for recoverable missing variables (\S\ref{sec:kg}).
  \item A 34K-sample graph-grounded finetuning dataset and LoRA training
        procedure for instilling three-action routing as a learned capability
        (\S\ref{sec:finetune}).
  \item A three-agent execution architecture with strict routing/generation
        separation, making planner and agent failures independently
        diagnosable (\S\ref{sec:agents}).

\end{itemize}

\paragraph{Paper organisation.}
Section~\ref{sec:related} surveys related work across the six threads that
PassiveQA draws from.  Section~\ref{sec:formalism} develops the formal
framework, including the information-state decomposition, utility-maximisation
framing, and the full set of retrieval and graph scoring signals.
Section~\ref{sec:data} describes dataset construction, balancing, and
variable population.  Section~\ref{sec:rag} presents the three RAG
architectures and their evaluation, establishing the inference-time ceiling.
Section~\ref{sec:kg} details the three-phase knowledge graph construction
and post-processing pipeline.  Section~\ref{sec:ft_data} describes the
finetuning dataset construction and prompt schema.
Section~\ref{sec:finetune} covers the LoRA training procedure.
Section~\ref{sec:agents} presents the three-agent execution architecture.
Section~\ref{sec:results} reports quantitative results, planner observations,
and qualitative error analysis.  Section~\ref{sec:limitations} discusses
limitations and future directions, and Section~\ref{sec:conclusion} concludes.
\begin{figure}[H]
\centering
\includegraphics[width=\textwidth]{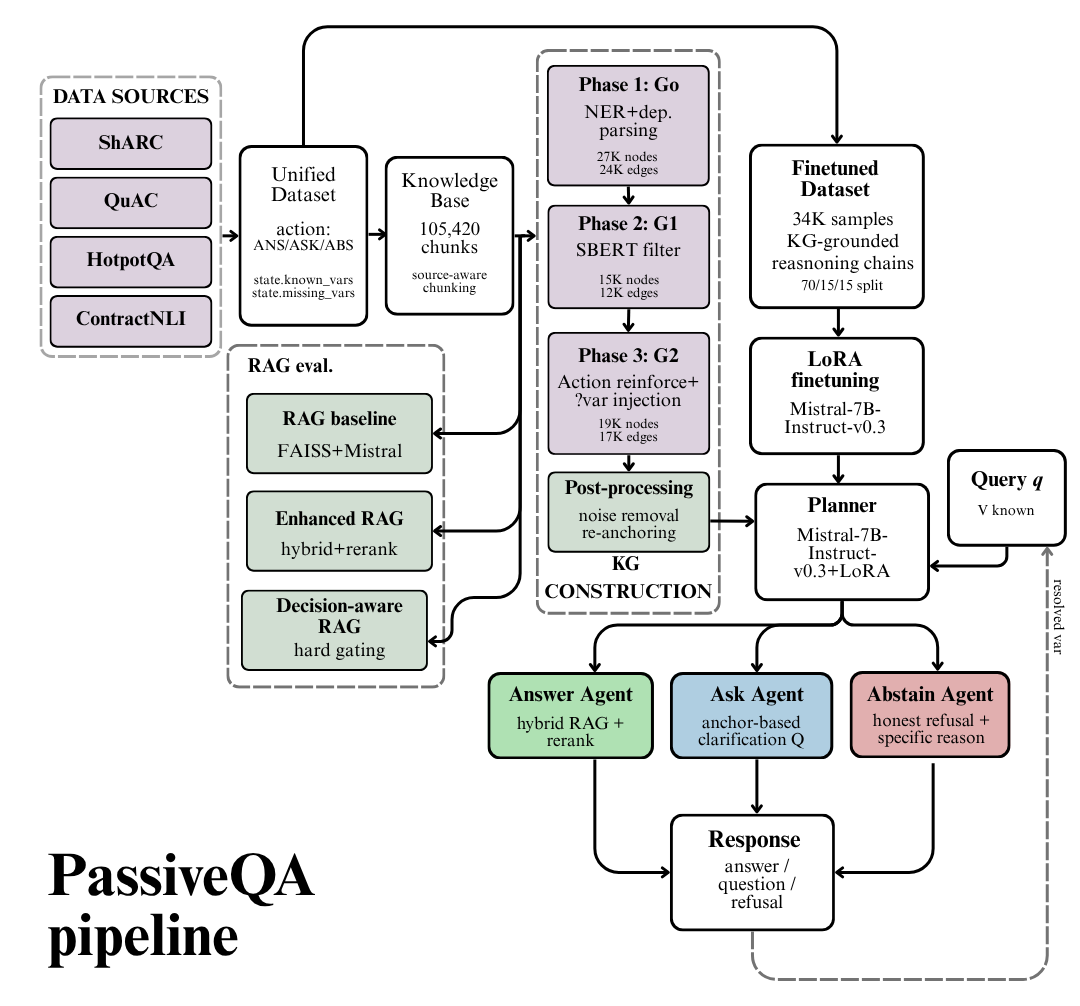} 
\caption{%
  Full PassiveQA pipeline. \textit{Left to right:} the four source datasets are merged
  into a unified 61K-sample schema with explicit variable-state fields (\S\ref{sec:data});
  a knowledge base of 105{,}420 chunks is constructed and indexed (\S\ref{sec:rag});
  three progressive RAG architectures are evaluated on the KB alone (\S\ref{sec:rag});
  the KB is simultaneously processed through a three-phase KG construction pipeline
  producing the decision-weighted graph $G_2$ (\S\ref{sec:kg}); $G_2$ and the unified
  dataset jointly generate the 34K KG-grounded finetuning dataset (\S\ref{sec:ft_data}),
  which trains the LoRA planner (\S\ref{sec:finetune}); at inference the planner
  receives the query and KG context and routes to one of three specialised agents
  (\S\ref{sec:agents}). The dashed feedback arrow models multi-turn state update
  (Eq.~\ref{eq:state_update}): a resolved variable from the \textsc{Ask} agent
  transitions from $V_{\mathrm{missing}}$ to $V_{\mathrm{known}}$ before the next
  planner call.
}
\label{fig:full_pipeline}
\end{figure}
\section{Related Work}
\label{sec:related}

\paragraph{Retrieval-Augmented Generation.}
RAG systems augment language models with non-parametric retrieval over an external
knowledge base~\citep{lewis2020rag,gao2023rag_survey}. Standard pipelines retrieve
top-$k$ chunks via dense passage retrieval~\citep{karpukhin2020dpr} or
retrieval-conditioned language modelling~\citep{guu2020realm}. Subsequent work explored
fusion-in-decoder generation~\citep{izacard2021fid} and retrieval from massive
datastores~\citep{borgeaud2022retro}. Hybrid pipelines combining
BM25~\citep{robertson2009bm25} with dense encoders and late-interaction
reranking~\citep{santhanam2022colbertv2} are now standard. Despite these advances,
retrieval improvements without explicit epistemic modelling tend to increase overconfident
answering — a finding we corroborate empirically in \S\ref{sec:rag}.

\paragraph{Hallucination and Faithfulness.}
Hallucination in LLMs is widely recognised as a critical obstacle to
deployment~\citep{ji2023hallucination,maynez2020faithfulness}.
Dedicated detection pipelines with calibrated ensembles have recently been proposed for
RAG settings~\citep{halt2025}, and selective abstention learning has been used as a
training-time mechanism to suppress hallucination at the
source~\citep{huang2025abstention}. Our work differs from all of the above by
targeting the \emph{decision to answer at all}, rather than post-hoc verification of a
generated answer~\citep{hallucination_mitigating_survey}.

\paragraph{Selective Prediction and Abstention.}
Selective prediction~\citep{geifman2017selective} studies the coverage--accuracy
trade-off. Xin et al.~\citep{xin2021abstention} formalise abstention for NLP via error
regularisation; Varshney and Baral~\citep{varshney2023post} extend this to open-domain
QA by revisiting abstained instances post-hoc. Recent empirical work shows that LLMs
frequently do not know when to refuse~\citep{yin2024llm_abstention}, and that confidence
entropy alone is an insufficient gating signal~\citep{phillips2026entropy}. We extend
selective prediction to a three-way open-ended QA routing decision.

\paragraph{Conversational and Clarification-seeking QA.}
The ShARC benchmark~\citep{saeidi2018sharc} introduced clarification-seeking before
committing to a policy-grounded answer, directly motivating our \textsc{Ask} action.
QuAC~\citep{choi2018quac} models information-seeking dialogues with explicit
unanswerable annotations. The importance of proactively asking clarifying questions was
established by Aliannejadi et al.~\citep{aliannejadi2019clarify}, and learning which
questions to generate was studied by Rao and Daumé~\citep{rao2018learning}. Recent work
on conversational machine reading formalises the decision to
inquire~\citep{zhou2023cmr}, but without a unified three-action framework or
graph-grounded supervision.
The original ShARC system achieved 61.4\% macro accuracy using task-specific
rule-following models~\citep{saeidi2018sharc}; PassiveQA does not target this
benchmark directly, as it is designed as a general-purpose epistemic routing
framework rather than a policy-compliance system, but the ShARC component of our
evaluation inherits its annotation scheme and supervision signal.

\paragraph{Knowledge Graphs for QA.}
Graph-augmented retrieval supports multi-hop reasoning by organising knowledge into
structured representations~\citep{survey_ras_2025}. QA-GNN~\citep{yasunaga2021qagnn}
jointly reasons over language models and knowledge graphs via graph neural networks;
Sun et al.~\citep{sun2019kgqa} ground multi-hop QA directly over entity subgraphs;
and KagNet~\citep{lin2019kagnet} integrates graph networks with commonsense
knowledge. Our work extends this paradigm by constructing a \emph{decision-reinforced}
knowledge graph whose edge weights encode the three-action supervision signal — a
dimension absent from prior graph-augmented QA systems.

\paragraph{Parameter-Efficient Finetuning and Alignment.}
LoRA~\citep{hu2021lora} and QLoRA~\citep{dettmers2023qlora} enable efficient
finetuning within single-GPU memory budgets. InstructGPT~\citep{ouyang2022rlhf}
demonstrated that RLHF can align model behaviour with human preferences, motivating
training-time alignment as the fundamental mechanism for behaviour change. Direct
Preference Optimisation~\citep{rafailov2023dpo} offers a stable supervised alternative.
We apply LoRA to instil three-action decision behaviour in
Mistral-7B-Instruct-v0.3~\citep{jiang2023mistral}, treating epistemic routing as a
behaviour alignment problem rather than a prompt engineering problem.

\section{Formal Framework}
\label{sec:formalism}

\subsection{Information State and Decision Policy}

Let $q$ denote a user query. We define the \emph{information state} of $q$ as:
\begin{equation}
  S(q) = \bigl(V_{\mathrm{known}},\; V_{\mathrm{missing}},\; C\bigr)
  \label{eq:state}
\end{equation}
where $V_{\mathrm{known}}$ is the set of concrete entities or attributes explicitly
present in the query and retrieved context, $V_{\mathrm{missing}}$ is the set of
variables required to resolve $q$ but absent from $S$, and $C$ is the set of
constraints and conversational context. The \emph{incompleteness measure} is:
\begin{equation}
  I(q) = \frac{|V_{\mathrm{missing}}|}{|V_{\mathrm{known}}| + |V_{\mathrm{missing}}|}
  \label{eq:incompleteness}
\end{equation}

The planner $\pi_\theta$ is a probabilistic policy over actions
$a \in \{\textsc{Answer}, \textsc{Ask}, \textsc{Abstain}\}$:
\begin{equation}
  \pi_\theta(a \mid S(q)) = \operatorname{softmax}\!\bigl(f_\theta(S(q))\bigr)
  \label{eq:policy}
\end{equation}
where $f_\theta$ is a neural network mapping the information state to action logits.

The planner's role is not to generate a response but to select the
action that maximises expected utility given $S(q)$. This separation
between \emph{decision} and \emph{generation} is the architectural
commitment that distinguishes PassiveQA from standard RAG systems,
which conflate the two into a single generation step.

\paragraph{State initialisation.}
At the start of each dialogue, the state is initialised from the
current query and retrieved context:
\begin{equation}
  S_0(q) = \bigl(\operatorname{ent}(q) \cup \operatorname{ent}(\mathcal{D}),\;
            \operatorname{req}(q) \setminus \operatorname{ent}(\mathcal{D}),\;
            \emptyset\bigr)
  \label{eq:state_init}
\end{equation}
where $\operatorname{ent}(\cdot)$ denotes named entities extracted from a
text and $\operatorname{req}(q)$ denotes the set of variables logically
required to answer $q$. The constraint set $C$ is empty at turn zero and
accumulates resolved variables and dialogue history across subsequent turns.

\paragraph{Action space semantics.}
The three actions partition the response space into non-overlapping
epistemic commitments. Formally, let $\mathcal{R}$ be the space of all
possible responses. The planner defines a routing function:
\begin{equation}
  \phi: S(q) \;\longrightarrow\;
  \begin{cases}
    \mathcal{R}_{\mathrm{ans}}   & a = \textsc{Answer}  \\
    \mathcal{R}_{\mathrm{clar}}  & a = \textsc{Ask}     \\
    \mathcal{R}_{\mathrm{ref}}   & a = \textsc{Abstain}
  \end{cases}
  \label{eq:routing}
\end{equation}
where $\mathcal{R}_{\mathrm{ans}}$ is the space of KB-grounded factual
responses, $\mathcal{R}_{\mathrm{clar}}$ is the space of well-formed
clarification questions, and $\mathcal{R}_{\mathrm{ref}}$ is the space
of specific, honest refusals. The routing function $\phi$ is implemented
by the finetuned planner at inference time and by the hard gate
(Eq.~\ref{eq:boundary}) in Architecture~3.

\paragraph{Relationship between incompleteness and action.}
The three actions correspond to three regimes of $I(q)$:
\begin{equation}
  a^*(q) \approx \begin{cases}
    \textsc{Answer}  & I(q) \approx 0 \\
    \textsc{Ask}     & 0 < I(q) < 1 \;\wedge\;
                       V_{\mathrm{missing}} \text{ is recoverable} \\
    \textsc{Abstain} & I(q) \approx 1 \;\vee\;
                       V_{\mathrm{missing}} \text{ is irrecoverable}
  \end{cases}
  \label{eq:action_regimes}
\end{equation}
The distinction between \textsc{Ask} and \textsc{Abstain} when
$V_{\mathrm{missing}} \neq \emptyset$ depends on whether the missing
information is \emph{recoverable} — i.e.\ whether a user clarification
could plausibly supply it. This recoverability judgement is encoded in
the knowledge graph via the presence or absence of \texttt{?var\_} nodes
(\S\ref{sec:kg}): a missing variable that can be injected as a graph
placeholder is recoverable; a query whose topic has no graph nodes at all
is not.

\subsection{Decision as Utility Maximisation}

Action selection is framed as expected utility maximisation:
\begin{equation}
  a^* = \arg\max_{a}\; \mathbb{E}\bigl[U(a \mid S(q))\bigr]
  \label{eq:utility}
\end{equation}
The utility function encodes the epistemic trade-offs:
\begin{itemize}[leftmargin=*]
  \item $U(\textsc{Answer})$: high reward for a correct, grounded response; high penalty for hallucination.
  \item $U(\textsc{Ask})$: moderate cost (extra conversational turn) but reduces $|V_{\mathrm{missing}}|$.
  \item $U(\textsc{Abstain})$: safe floor; avoids harm but provides no useful information.
\end{itemize}

\subsection{Hallucination Risk}

We define hallucination risk as the conditional probability of generating an incorrect answer
when committing to \textsc{Answer}:
\begin{equation}
  H(q) = P\!\bigl(\text{incorrect answer} \mid S(q),\; a = \textsc{Answer}\bigr)
  \label{eq:hallucination}
\end{equation}
We hypothesise a structural relationship between hallucination risk and information incompleteness:
\begin{equation}
  H(q) \propto I(q)
  \label{eq:hall_prop}
\end{equation}
\emph{Hallucination risk increases monotonically with information incompleteness.}
This motivates the use of $I(q)$ as a gating signal before answer generation. This relationship is motivated by observed trends in our experiments, though we do not provide explicit statistical validation.

\subsection{Retrieval Signals}

Two complementary retrieval-derived signals are formalised. \emph{Confidence} measures
the best-matching chunk in the document set $\mathcal{D}$:
\begin{equation}
  \operatorname{Conf}(q) = \max_{d \in \mathcal{D}}\; \operatorname{sim}(q, d)
  \label{eq:conf}
\end{equation}
\emph{Coverage} measures lexical completeness of the query terms against the retrieved set:
\begin{equation}
  \operatorname{Coverage}(q) = \frac{|\operatorname{terms}(q) \cap \operatorname{terms}(\mathcal{D})|}{|\operatorname{terms}(q)|}
  \label{eq:coverage}
\end{equation}

\subsection{Ambiguity Score}

The ambiguity signal is a feature-based estimator over $K$ heuristic signals
$h_i(q) \in [0,1]$:
\begin{equation}
  \operatorname{Amb}(q) = \frac{1}{K}\sum_{i=1}^{K} h_i(q)
  \label{eq:ambiguity}
\end{equation}
The heuristics include: query length ($\leq 4$ words), presence of dangling pronouns,
vague quantifiers, absence of named entities, and comparative constructions without
both comparison arguments.

\subsection{Conflict Score}

When multiple retrieved chunks address the same query from contradictory
directions, committing to \textsc{Answer} risks producing an internally
inconsistent response. We define the \emph{conflict score} as the mean
pairwise cosine dissimilarity among the top-$k$ retrieved chunk embeddings
$\{e_1, \ldots, e_k\}$:
\begin{equation}
  \operatorname{Conflict}(q) = 1 - \frac{2}{k(k-1)}
  \sum_{i=1}^{k-1}\sum_{j=i+1}^{k}
  \operatorname{sim}(e_i, e_j)
  \label{eq:conflict}
\end{equation}
A high conflict score ($\operatorname{Conflict}(q) > \tau_{\mathrm{con}}$)
indicates that the retrieved evidence is internally inconsistent — the query
may be ambiguous across document boundaries, or the KB contains contradictory
entries. In either case, the appropriate response is to seek clarification
rather than to synthesise a potentially misleading answer. Conflict thus
serves as the highest-priority signal in the hard gating rule
(Eq.~\ref{eq:boundary}).

\subsection{Joint Answerability Signal}

The four signals — $\operatorname{Conf}(q)$, $\operatorname{Coverage}(q)$,
$\operatorname{Amb}(q)$, and $\operatorname{Conflict}(q)$ — are each
necessary but not individually sufficient for routing. We define a scalar
\emph{joint answerability signal} $A(q) \in [0, 1]$ that summarises
evidential support:
\begin{equation}
  A(q) = \operatorname{Conf}(q) \cdot \operatorname{Coverage}(q)
         \cdot \bigl(1 - \operatorname{Amb}(q)\bigr)
         \cdot \bigl(1 - \operatorname{Conflict}(q)\bigr)
  \label{eq:answerability}
\end{equation}
Each factor lies in $[0,1]$, so $A(q) = 1$ only when retrieval confidence
is maximal, query terms are fully covered, the query is unambiguous, and
retrieved chunks are mutually consistent — the ideal conditions for
\textsc{Answer}. When any factor approaches zero, $A(q)$ collapses
regardless of the others, encoding the intuition that a single failure mode
is sufficient to warrant caution. The hard gate (Eq.~\ref{eq:boundary})
can equivalently be read as a threshold rule on Eq.~\ref{eq:answerability}
with per-factor thresholds that allow individual signals to override the
product when their deviation is extreme.

\subsection{Variable Resolution and Information Gain}

In multi-turn settings, the \textsc{Ask} action is only valuable if it
reduces incompleteness. We define the \emph{resolution rate} at turn $t$ as:
\begin{equation}
  \rho_t = \frac{|V_{\mathrm{known},t}|}{|V_{\mathrm{known},t}|
            + |V_{\mathrm{missing},t}|}
            = 1 - I_t(q)
  \label{eq:resolution}
\end{equation}
A successful \textsc{Ask} at turn $t$ transitions one or more variables from
$V_{\mathrm{missing},t}$ to $V_{\mathrm{known},t+1}$ per
Eq.~\ref{eq:state_update}, strictly increasing $\rho_{t+1} > \rho_t$.
The expected information gain of issuing a clarification question $c$ is:
\begin{equation}
  \mathrm{IG}(c \mid S_t) = \mathbb{E}\!\left[
    \rho_{t+1} - \rho_t \;\middle|\; S_t,\; a_t = \textsc{Ask},\; c
  \right]
  \label{eq:infogain}
\end{equation}
A well-formed clarification question maximises $\mathrm{IG}$ by targeting
the specific missing variable that is both (a) necessary to resolve the
query and (b) most likely to be known to the user. The anchor-based question
builder (\S\ref{sec:ft_data}) implements a heuristic approximation of this
objective: it selects the missing variable with the highest SBERT similarity
to the known variable set, targeting the gap most proximate to information
the user has already provided.

\subsection{Knowledge Graph Scoring}

Each edge $e$ in the knowledge graph receives a final weight combining semantic
grounding (Phase~2) and query-action reinforcement (Phase~3):
\begin{equation}
  w_e = \alpha \cdot \operatorname{sem}(e) + (1 - \alpha) \cdot \operatorname{act}(e)
  \label{eq:edge_weight}
\end{equation}
where $\operatorname{sem}(e)$ is the semantic similarity score between the triple
and its source passage, and $\operatorname{act}(e)$ is the accumulated action
reinforcement signal. With $\alpha = 0.5$, both components receive equal weight.

The path score over a graph path $p$ in the set of all paths $\mathcal{P}(q)$ is:
\begin{equation}
  \operatorname{PathScore}(q) = \max_{p \in \mathcal{P}(q)} \prod_{e \in p} w_e
  \label{eq:path_score}
\end{equation}
Strong paths ($\operatorname{PathScore} \approx 1$) support \textsc{Answer};
broken paths support \textsc{Ask}; absent paths support \textsc{Abstain}.

\subsection{Training Objective}

The finetuning loss decomposes into a decision classification component and a generation
component:
\begin{equation}
  \mathcal{L} = \mathcal{L}_{\mathrm{decision}} + \lambda\, \mathcal{L}_{\mathrm{generation}}
  \label{eq:loss}
\end{equation}
where $\mathcal{L}_{\mathrm{decision}} = -\sum_{(q,a,y)} \log P_\theta(a \mid S(q))$
is the cross-entropy loss over action labels and $\mathcal{L}_{\mathrm{generation}}$
is the standard causal language modelling loss over the structured reasoning chain.

\subsection{Multi-turn State Update}

For multi-turn dialogues, the information state is updated after each turn:
\begin{equation}
  S_{t+1} = S_t \cup \operatorname{resolve}(q_t, a_t)
  \label{eq:state_update}
\end{equation}
where $\operatorname{resolve}$ transfers variables from $V_{\mathrm{missing}}$ to
$V_{\mathrm{known}}$ when the user's response satisfies a clarification request.

\subsection{Decision Boundary}

The hard gating rule in Architecture~3 implements the following decision boundary:
\begin{equation}
  a = \begin{cases}
    \textsc{Answer}  & \text{if } \operatorname{Conf}(q) > \tau_c \;\wedge\; I(q) \approx 0 \\[4pt]
    \textsc{Ask}     & \text{if } I(q) > \tau_i \;\vee\; \operatorname{Amb}(q) > \tau_a \\[4pt]
    \textsc{Abstain} & \text{otherwise}
  \end{cases}
  \label{eq:boundary}
\end{equation}
where $\tau_c$, $\tau_i$, $\tau_a$ are tunable thresholds.

\subsection{Relationship Between Formal Signals and Implementation}

The equations in this section range from directly implemented to
approximately realised. Table~\ref{tab:theory_impl} summarises the mapping.

\begin{table}[ht]
\centering
\caption{Mapping from formal signals to implementation components.}
\label{tab:theory_impl}
\begin{tabular}{llp{5.5cm}}
\toprule
\textbf{Equation} & \textbf{Status} & \textbf{Implementation} \\
\midrule
$I(q)$ (Eq.~\ref{eq:incompleteness})
  & Direct & Variable population via GPT-4o-mini (\S\ref{sec:data});
    threshold $\tau_i$ in hard gate (Eq.~\ref{eq:boundary}) \\
$\operatorname{Conf}(q)$ (Eq.~\ref{eq:conf})
  & Direct & Max sigmoid cross-encoder logit, Architecture~3 \\
$\operatorname{Coverage}(q)$ (Eq.~\ref{eq:coverage})
  & Direct & Non-stopword term overlap, Architecture~3 \\
$\operatorname{Amb}(q)$ (Eq.~\ref{eq:ambiguity})
  & Direct & Five heuristic signals, Architecture~3 \\
$\operatorname{Conflict}(q)$ (Eq.~\ref{eq:conflict})
  & Direct & Pairwise chunk dissimilarity, Architecture~3 \\
$A(q)$ (Eq.~\ref{eq:answerability})
  & Approximated & The hard gate uses per-signal thresholds rather than
    the product form; $A(q)$ provides the theoretical motivation
    for why all four signals are jointly necessary \\
$\operatorname{PathScore}(q)$ (Eq.~\ref{eq:path_score})
  & Approximated & Edge weights are used to rank triples during KG
    context retrieval (\S\ref{sec:ft_data}); explicit path
    enumeration is not computed at inference time \\
$\mathrm{IG}(c \mid S_t)$ (Eq.~\ref{eq:infogain})
  & Heuristic & The anchor-based question builder (\S\ref{sec:ft_data})
    maximises SBERT similarity between missing variable and known
    variable set as a proxy; no empirical validation is conducted \\
$\mathcal{L}$ (Eq.~\ref{eq:loss})
  & Direct & Causal LM loss over assistant turn tokens, $\lambda=1$ \\
\bottomrule
\end{tabular}
\end{table}

Equations $A(q)$, $\operatorname{PathScore}(q)$, and $\mathrm{IG}(c \mid S_t)$
are best understood as conceptual abstractions of the decision process, with current implementations relying on heuristic approximations. Grounding these signals in learnable parameters
— for instance, learning $\alpha$ in Eq.~\ref{eq:edge_weight} end-to-end, or
training a dedicated path scorer — is deferred to future work.

\section{Dataset Construction}
\label{sec:data}

\subsection{Dataset Selection}

Four publicly available datasets were selected to provide complementary coverage of the
three target actions across diverse reasoning contexts.

\paragraph{ShARC~\citep{saeidi2018sharc}.}
A multi-turn dialogue dataset grounded in government policy documents. Each sample
contains a regulatory snippet, an optional user scenario, a question, and a gold label.
The label space maps directly onto the three actions: Yes/No responses map to
\textsc{Answer} (13{,}830 samples); Follow-on responses — where the system must request
additional information — map to \textsc{Ask}; and Irrelevant labels map to
\textsc{Abstain} (1{,}256 samples). ShARC provides explicit evidence chains (clarification
questions) that serve as primary supervision for the \textsc{Ask} action.
Approximately 81.6\% of samples include a user scenario, and dialogue history depth
ranges from 0 to 6 prior turns.

\paragraph{QuAC~\citep{choi2018quac}.}
A large-scale conversational reading comprehension dataset built on Wikipedia passages,
comprising 83{,}568 training QA pairs across 11{,}567 dialogues (avg.\ 7.2 turns).
\texttt{CANNOTANSWER} responses (17.3\% of training samples) are mapped to
\textsc{Abstain}; turns marked \texttt{followup=y} are mapped to \textsc{Ask};
all remaining answered turns map to \textsc{Answer}. QuAC is the primary source
of multi-turn \textsc{Answer} supervision.

\paragraph{HotpotQA~\citep{yang2018hotpotqa}.}
A multi-hop QA dataset with 90{,}447 training samples requiring reasoning over two or
more Wikipedia passages. All samples have gold answers; the entire split maps to
\textsc{Answer}. The dataset contributes two reasoning types: \emph{bridge} questions
(72{,}991 samples) requiring chained inference across documents, and \emph{comparison}
questions (17{,}456 samples) requiring attribute comparison.

\paragraph{ContractNLI~\citep{koreeda2021contractnli}.}
Legal contract review framed as natural language inference over 423 NDA documents,
with 7{,}191 clause-level annotations across 17 NDA clause types. Entailment and
Contradiction labels both map to \textsc{Answer} (60.8\%); NotMentioned maps to
\textsc{Abstain} (39.2\%). This dataset provides a controlled \textsc{Abstain} source
where abstention arises from genuine absence of relevant information rather than
query-context mismatch.

\subsection{Unified Schema}

All four datasets are converted into a single JSON schema with fields for \texttt{query},
\texttt{context\_documents}, \texttt{action\_label}, \texttt{response}, a structured
\texttt{state} object (\texttt{known\_variables}, \texttt{missing\_variables},
\texttt{failure\_mode}, \texttt{completeness}), and \texttt{metadata}
(source, multi-turn flag, turn ID, difficulty).

The ShARC evidence chain directly populates \texttt{missing\_variables}, providing explicit
\textsc{Ask} supervision. The final merged training set comprises 203{,}096 samples with
the distribution: \textsc{Answer} 65.1\%, \textsc{Ask} 22.4\%, \textsc{Abstain} 12.5\%.

\subsection{Dataset Balancing}

The raw merged dataset is heavily skewed toward \textsc{Answer} (65.1\%), which would
bias the planner toward reflexive answering — precisely the behaviour this work aims
to counteract. A balanced 61K subset was constructed with the target distribution:
\textsc{Answer}~33\% $\cdot$ \textsc{Ask}~37\% $\cdot$ \textsc{Abstain}~30\%.

Dialogues were treated as atomic units during sampling to preserve conversational
coherence. Turn depth was capped per dialogue length. Of 12{,}195 unique dialogues, 609
were capped; all resulting sequences were verified for continuity with zero broken chains.

Source minimums were enforced: ContractNLI (7K — all available), HotpotQA (14K),
ShARC (14K), QuAC ($\sim$26K). The achieved action distribution after trimming settled at
\textsc{Answer}~49.9\% / \textsc{Ask}~28.9\% / \textsc{Abstain}~21.2\%, shifted
substantially toward passive actions. The hard \textsc{Answer} floor from HotpotQA
(which contributes no \textsc{Ask}/\textsc{Abstain} signal) prevented reaching the
original targets exactly.

\subsection{Variable Population via GPT-4o-mini}

The unified schema includes \texttt{known\_variables} and \texttt{missing\_variables}
fields. For ShARC, missing variables were directly available from the evidence chain.
For all other sources, these fields were populated via GPT-4o-mini under controlled prompting. While this introduces potential annotation noise, the variables serve as weak supervision signals, and the model primarily learns aggregate decision patterns rather than relying on exact variable identity. The extraction prompt instructed the model to identify:
(a)~\texttt{known\_variables} — concrete entities or attributes explicitly present in
the query (max~5); and (b)~\texttt{missing\_variables} — information required to resolve
the query but absent, with the hard constraint that \textsc{Answer}-action samples always
receive an empty \texttt{missing\_variables} list.

Extraction was parallelised in batches of 50 with 5 concurrent workers and
checkpoint-based crash recovery. The resulting variable fields serve dual purposes:
inputs to the knowledge graph builder (known variables anchor graph traversal) and
the core supervision signal for the planner's finetuning dataset.

Table~\ref{tab:dataset_summary} summarises the final dataset composition.

\begin{table}[ht]
\centering
\caption{Final merged dataset composition across sources and action labels.}
\label{tab:dataset_summary}
\begin{tabular}{lrrrr}
\toprule
\textbf{Source} & \textbf{Total} & \textbf{Answer} & \textbf{Ask} & \textbf{Abstain} \\
\midrule
ShARC       & 14{,}000 & -- & Primary & 1{,}256 \\
QuAC        & 26{,}000 & Primary & Followup & \texttt{CANNOTANSWER} \\
HotpotQA    & 14{,}000 & All & -- & -- \\
ContractNLI &  7{,}000 & 60.8\% & -- & 39.2\% \\
\midrule
\textbf{Merged} & \textbf{61K} & 49.9\% & 28.9\% & 21.2\% \\
\bottomrule
\end{tabular}
\end{table}

\begin{table}[ht]
\centering
\caption{Dataset characteristics across the four sources.}
\label{tab:dataset_eda}
\begin{tabular}{lrrrr}
\toprule
\textbf{Property} & \textbf{ShARC} & \textbf{QuAC} & \textbf{HotpotQA} & \textbf{ContractNLI} \\
\midrule
Avg.\ question length (words)  & 7.5  & 6.5   & 17.8    & 18.2  \\
Avg.\ context length (words)   & 45.5 & 400+  & varies  & 1{,}674 \\
Multi-turn                     & Yes (up to 6) & Yes (avg 7.2) & No & No \\
\textsc{Ask} signal source     & Evidence chain & \texttt{followup=y} & None & None \\
\textsc{Abstain} signal source & Irrelevant label & \texttt{CANNOTANSWER} & None & NotMentioned \\
Reasoning type                 & Policy & Wikipedia & Multi-hop & NLI \\
Avg.\ turns per dialogue       & 2.3  & 7.2   & 1.0     & 1.0   \\
\% queries with named entities & 61.2 & 74.8  & 92.1    & 88.4  \\
\bottomrule
\end{tabular}
\end{table}

\noindent The contrast in context length between ShARC (45.5 words — compact policy
snippets) and ContractNLI (1{,}674 words — full NDA documents) motivates the
source-aware chunking strategy described in \S\ref{sec:rag}: uniform chunking would
either fragment ShARC snippets below minimum coherence or leave ContractNLI documents
unchunked and oversized for retrieval.

\section{RAG Architectures}
\label{sec:rag}

We evaluate three progressively more sophisticated RAG architectures on a balanced
900-sample held-out set (300 per action) before introducing the graph-grounded planner.

\subsection{Architecture 1 — Baseline RAG}

The baseline follows a standard retrieve-then-generate design. Context documents are
deduplicated by MD5 hash and indexed with source-aware chunking: ContractNLI documents
(dense legal text, typically 1{,}600+ words) are chunked when exceeding 300 words;
QuAC passages when exceeding 400 words; ShARC and HotpotQA are kept whole. A
fixed word-window strategy with 50-word overlap is used.

All chunks are encoded with \texttt{all-MiniLM-L6-v2} (384-dim) and indexed in a FAISS
flat inner-product index over L2-normalised vectors for exact cosine similarity search.
At inference, the top-$k$ chunks are retrieved, concatenated with the query and
conversation history, and passed to Mistral-7B-Instruct-v0.3 (4-bit quantised, NF4)
with a system prompt enforcing the three-action output format. Decoding is greedy
(temperature=1.0, do\_sample=False).

\paragraph{Results.} Decision accuracy: 34\%; hallucination rate: 42.7\%; macro F1: 26.7\%.
Per-action accuracy reveals severe skew: \textsc{Answer} predicted correctly 81\% of the
time; \textsc{Ask} accuracy collapses to 12\%; \textsc{Abstain} to 9\%. The confusion
matrix shows the model defaulting overwhelmingly to \textsc{Answer}: 80 of 100 \textsc{Abstain}
cases and 75 of 100 \textsc{Ask} cases are incorrectly classified as \textsc{Answer}.
This provides strong support for the central thesis: a standard RAG system with no epistemic passivity training
is effectively incapable of distinguishing when to ask or abstain, reducing to a
near-unconditional answering system.

\subsection{Architecture 2 — Enhanced RAG}

The enhanced pipeline introduces five targeted modifications addressing specific failure
modes identified in the baseline.

\paragraph{Multi-Granularity Knowledge Base.}
The KB is indexed at two granularities: coarse (full semantic chunks via sentence-boundary-aware
chunking) and fine (individual sentences). Both are embedded in a single FAISS index.

\paragraph{Hybrid Retrieval.}
A BM25 sparse index is added alongside the dense index. At query time, scores from
both are normalised to $[0,1]$ and fused as a weighted sum with $\alpha = 0.5$,
addressing vocabulary mismatch for domain-specific regulatory and legal text.

\paragraph{Query Understanding.}
Two pre-retrieval transformations are applied: (i) query rewriting via Mistral to produce
retrieval-friendly formulations of elliptical multi-turn queries; and (ii) multi-hop
detection that decomposes queries likely requiring multi-document evidence into 2--3
independent sub-queries.

\paragraph{Cross-Encoder Reranking and Context Compression.}
The top-20 hybrid results are reranked using \texttt{cross-encoder/ms-marco-MiniLM-L-6-v2}.
The top-5 chunks are then compressed to retain only query-relevant sentences via
keyword overlap scoring.

\paragraph{Self-Reflection.}
A second prompt asks the model to verify its action decision. Reflection changed the
initial decision in 12.7\% of cases .

\paragraph{Results.} Despite these five enhancements, overall decision accuracy remains
at 34\% with macro F1 of 26.7\% — statistically identical to the baseline. Per-action
accuracy is likewise unchanged: \textsc{Answer} 81\%, \textsc{Ask} 12\%, \textsc{Abstain}
9\%. The confusion structure is preserved verbatim from the baseline.
Hallucination rate rose to 51.7\%, demonstrating that better retrieval without a decision gate
can \emph{increase} overconfident answering by surfacing more plausible-looking but
insufficient evidence. This consistent negative result highlights a key limitation of the RAG section:
the \textsc{Answer}-skew failure mode is unlikely to be solely a retrieval quality problem, and not easily addressed through inference-time modifications alone.
The root cause lies in the model's underlying generation priors — the LLM has no finetuned
disposition toward epistemic passivity.

\subsection{Architecture 3 — Decision-Aware RAG (v3)}

Architecture~3 abandons the single-prompt generate-and-decide paradigm. Instead, a
dedicated pre-generation pipeline of explicit, interpretable signals gates the LLM
before any response is generated.

\paragraph{Evidence Scoring.}
Three signals are computed directly from retrieved chunks before any generation call.
\emph{Confidence} uses the maximum sigmoid-normalised cross-encoder logit score.
\emph{Coverage} measures the fraction of non-stopword query terms appearing across
the retrieved chunk set.
\emph{Ambiguity} is computed via Equation~\eqref{eq:ambiguity} over five heuristic signals.
\emph{Conflict} measures pairwise cosine dissimilarity among the top-4 chunk embeddings.

\paragraph{Answerability Classifier.}
A dedicated lightweight LLM call is issued before answer generation, asking only
whether the context is sufficient. The classifier outputs one of three labels:
\texttt{ANSWERABLE}, \texttt{NEEDS\_CLARIFICATION}, or \texttt{NOT\_ANSWERABLE},
decoupling epistemic judgement from fluent answer generation.

\paragraph{Hard Gating.}
The hard gate combines all signals with explicit priority ordering:
\begin{enumerate}[leftmargin=*, label=\arabic*.]
  \item $\operatorname{Conflict} > 0.70 \;\Rightarrow\; \textsc{Ask}$
  \item $\operatorname{Conf}(q) < 0.35 \;\wedge\; \operatorname{Coverage}(q) < 0.30 \;\Rightarrow\; \textsc{Abstain}$
  \item $\operatorname{Amb}(q) > 0.45 \;\Rightarrow\; \textsc{Ask}$
  \item Classifier $= \texttt{NOT\_ANSWERABLE} \;\Rightarrow\; \textsc{Abstain}$
  \item Classifier $= \texttt{NEEDS\_CLARIFICATION} \;\Rightarrow\; \textsc{Ask}$
  \item None of the above $\;\Rightarrow\; \textsc{Answer}$
\end{enumerate}
These thresholds are tunable without retraining, making this the only component where
passive-action bias is directly controllable.

\paragraph{Action-Specific Generation.}
Three separate task-specific prompts handle the three actions. The \textsc{Ask} prompt
receives the gate reason as explicit input and enforces a single well-formed clarification
question. The \textsc{Abstain} prompt generates a specific statement about missing
information rather than a generic refusal.

\paragraph{Results.} Architecture~3 is the only pipeline achieving measurable
improvement: decision accuracy 38\%, macro F1 35.3\% (+8.6 pp), hallucination rate
33.8\% ($-$9 pp), and \textsc{Ask} recall 40\% (vs.\ 2\% baseline). However,
\textsc{Abstain} recall remains low at 13.3\%, and all architectures expose
the ceiling of inference-time interventions, motivating the finetuning approach.

\subsection{Quantitative Comparison}

\begin{table}[ht]
\centering
\caption{Quantitative comparison of three RAG architectures on the 900-sample evaluation set.}
\label{tab:rag_results}
\begin{tabular}{lrrr}
\toprule
\textbf{Metric}            & \textbf{Baseline} & \textbf{Enhanced} & \textbf{v3} \\
\midrule
Decision Accuracy          & 34.0\%            & 34.0\%            & 38.0\%      \\
Macro F1                   & 26.7\%            & 26.7\%            & 35.3\%      \\
Hallucination Rate         & 42.7\%            & 51.7\%            & 33.8\%      \\
\textsc{Ask} Recall        &  2.0\%            & 12.0\%            & 40.0\%      \\
\textsc{Abstain} Recall    & 26.0\%            &  9.0\%            & 13.3\%      \\
Coverage                   & 67.3\%            & 78.7\%            & 54.0\%      \\
\bottomrule
\end{tabular}
\end{table}

\begin{figure}[htbp]
\centering
\resizebox{\textwidth}{!}{%
\begin{tikzpicture}[
  node distance=0.45cm and 0.85cm,
  block/.style={rectangle, rounded corners=2pt, draw=black, thick,
                fill=lightgray, text width=2.2cm, align=center,
                minimum height=0.75cm, font=\small},
  decision/.style={diamond, draw=black, thick, fill=lightgray,
                   aspect=2.2, align=center, font=\footnotesize,
                   inner sep=1pt},
  arrow/.style={-{Stealth[length=4pt]}, thick},
  dasharrow/.style={-{Stealth[length=4pt]}, thick, dashed}
]
  \node[block] (q)     {Query $q$};
  \node[block, right=of q]   (ret)   {Hybrid\\Retrieval};
  \node[block, right=of ret] (score) {Evidence\\Scoring};
  \node[decision, right=0.9cm of score] (gate) {Hard\\Gate};

  \node[block, above right=0.65cm and 0.75cm of gate] (ans) {\textsc{Answer}\\Agent};
  \node[block, right=1.5cm of gate]                   (ask) {\textsc{Ask}\\Agent};
  \node[block, below right=0.65cm and 0.75cm of gate] (abs) {\textsc{Abstain}\\Agent};

  \draw[arrow] (q)     -- (ret);
  \draw[arrow] (ret)   -- (score);
  \draw[arrow] (score) -- (gate);
  \draw[arrow] (gate.north east) -- node[above,sloped,font=\scriptsize]{Answer}  (ans.west);
  \draw[arrow] (gate.east)       -- node[above,font=\scriptsize]{Ask}            (ask.west);
  \draw[arrow] (gate.south east) -- node[below,sloped,font=\scriptsize]{Abstain} (abs.west);

  \node[block, above=1.1cm of score, text width=3.0cm] (planner)
    {Finetuned Planner\\[-1pt]\small(Mistral-7B-Instruct-v0.3 + LoRA)};
  \draw[dasharrow] (score.north) -- (planner.south)
    node[midway,right,font=\scriptsize]{signals};
  \draw[dasharrow] (planner.east) -| (gate.north)
    node[near start,above,font=\scriptsize]{decision};

\end{tikzpicture}%
}
\caption{%
  Overview of the PassiveQA pipeline. After hybrid retrieval and evidence scoring,
  either the hard gate (Architecture~3) or the finetuned planner routes the query to one
  of three specialised agents. Dashed arrows indicate the planner path used in the
  full three-agent architecture.
}
\label{fig:pipeline}
\end{figure}
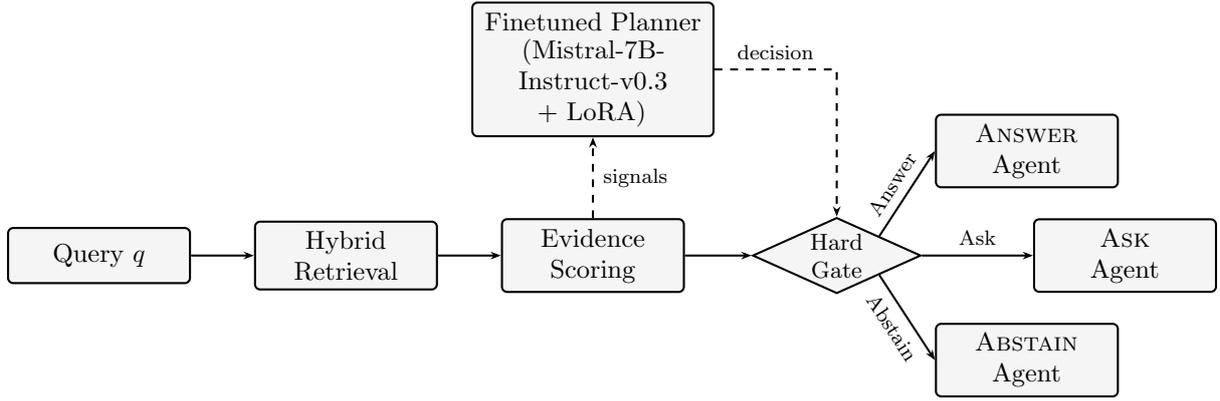

\section{Knowledge Graph Construction}
\label{sec:kg}

The knowledge graph is built through a three-phase pipeline that transforms flat KB chunks
into a structured, decision-weighted graph. The key novelty is that the graph is not a
passive factual store — it is actively shaped by the three-action supervision signal, so
edge weights encode both semantic coherence and epistemic utility.

\subsection{Phase 1 — Entity and Relation Extraction ($G_0$)}

All 105{,}420 KB chunks are parsed using spaCy (\texttt{en\_core\_web\_sm}). Two
design constraints distinguish this extraction from standard open information extraction.

\emph{Named-entity-only nodes.} Only spaCy NER entities are admitted as graph nodes;
noun chunks are excluded to avoid high-degree hub nodes from generic phrases. Entities
are normalised, filtered through a hard validator (rejecting pronouns, stopwords,
pure numerics, strings under 3 characters), and mapped to seven semantic categories:
Person, Organisation, Location, Attribute, Work, Concept, Event.

\emph{Entity-constrained triples.} Relations are extracted via syntactic dependency
parsing. A triple is admitted only if both subject and object map to named entities
in the entity set, eliminating pronoun-based spurious hub nodes. Weak verbs (copulae,
auxiliaries) are filtered except in HotpotQA, where relational verbs carry genuine
multi-hop signal. Initial confidence scores: $0.8$ for direct SVO triples, $0.7$ for
prepositional-object chains.

$G_0$ after Phase~1: 27{,}189 nodes, 24{,}491 edges. Isolated nodes are immediately pruned.

\subsection{Phase 2 — Semantic Validation ($G_1$)}

Each triple is converted to a natural-language sentence via a template. Both KB chunk
texts and triple sentences are encoded with \texttt{all-MiniLM-L6-v2}. For each edge,
the cosine similarity between the triple sentence embedding and its source chunk embeddings
measures semantic grounding. A frequency bonus $\log(1 + \text{freq}) \times 0.03$
rewards cross-document corroboration. Edges below $\tau = 0.50$ are removed.

$G_1$ after Phase~2: 15{,}468 nodes, 11{,}965 edges — 12{,}526 noisy edges removed
(51.1\% of $G_0$ edges). Average edge weight rises slightly from 0.726 to 0.728,
confirming that removed edges were low-confidence extractions.

\subsection{Phase 3 — Query-Guided Decision Reinforcement ($G_2$)}

Phase~3 is the central novel contribution of the KG module. For each of the 273{,}809
(query, KB, action) triples in the dataset, the associated graph edges are updated
according to the gold action label:
\begin{itemize}[leftmargin=*]
  \item \textsc{Answer}: edge weight $+0.20$ (context was sufficient to answer).
  \item \textsc{Ask}: edge weight $+0.05$ (context partially relevant but insufficient).
  \item \textsc{Abstain}: edge weight $-0.10$ (context despite surface relevance did not support answering).
\end{itemize}
For HotpotQA entries, the shortest path between pairs of KB-linked nodes is found and
each edge along that path is reinforced, propagating multi-hop signal through the graph.

\emph{Variable node injection.} For every \textsc{Ask}-action sample, query entities
absent from the graph are added as special \texttt{?var\_} nodes connected to the nearest
graph entity via a \texttt{requires} relation with weight~0.9. These nodes serve as
structural placeholders representing missing information. A total of 4{,}295 variable
nodes were injected (21.7\% of total nodes).

To illustrate the injection mechanism concretely: \texttt{?var\_quac\_000014}, anchored
to \emph{the United States}, encodes the missing entity in the query \emph{``What do we
know about Cove Reber?''}; while \texttt{?var\_quac\_000051}, anchored to \emph{the Air
Force}, encodes the unresolved gap in \emph{``Where was Bernie born?''} — each
representing the precise information the \textsc{Ask} agent should surface to the user.

The final edge weight combines both signals per Equation~\eqref{eq:edge_weight}.

$G_2$ after Phase~3: 19{,}763 nodes (4{,}295 variable), 17{,}715 edges.
Average edge weight rises to 0.819, reflecting reinforcement of frequently traversed
\textsc{Answer} paths.

\subsection{Graph Statistics}

Table~\ref{tab:kg_stats} summarises the knowledge graph at each construction phase.

\begin{table}[ht]
\centering
\caption{Knowledge graph statistics across construction phases.}
\label{tab:kg_stats}
\begin{tabular}{lrrrr}
\toprule
\textbf{Phase} & \textbf{Nodes} & \textbf{Edges} & \textbf{Avg.\ Weight} & \textbf{Notes} \\
\midrule
$G_0$ & 27{,}189 & 24{,}491 & 0.726 & Raw extraction \\
$G_1$ & 15{,}468 & 11{,}965 & 0.728 & Semantic validation \\
$G_2$ & 19{,}763 & 17{,}715 & 0.819 & Incl.\ 4{,}295 variable nodes \\
\bottomrule
\end{tabular}
\end{table}

The entity distribution across the 15{,}468 real nodes: Person (31.2\%), Organisation (26.8\%),
Attribute (18.1\%), Location (11.9\%), Work (7.7\%), Concept (2.4\%), Event (1.9\%).
The graph is deliberately sparse: average real node degree 1.55, organised into 3{,}907
weakly connected components, the largest spanning 5{,}834 nodes. Edge weight distribution
shows 47.1\% of edges in the top bucket (0.8--1.0), a direct consequence of Phase~3
reinforcement concentrating weight on \textsc{Answer}-supporting paths. Top relation
types are semantically rich and source-appropriate: legal relations from ContractNLI
(\texttt{disclose\_to}, \texttt{transfer\_from}), biographical relations from QuAC
(\texttt{bear\_in}, \texttt{refer\_as}), and event relations from HotpotQA
(\texttt{defeat}, \texttt{win}, \texttt{release\_on}).

\subsection{Graph Post-Processing}

Following Phase~3, the graph undergoes a structured post-processing pass that repairs
three categories of quality issues without rerunning any construction phase.

\paragraph{Noise node removal.}
Despite the hard validator in Phase~1, a residual population of low-quality nodes
survives into $G_2$: (i)~mixed alphanumeric strings that passed the length filter
(e.g.\ \emph{the 1980s}); (ii)~short non-entity tokens marginally exceeding the
three-character threshold; (iii)~generic hub nodes with degree exceeding three times
the graph mean whose text spaCy does not recognise as any named entity type; and
(iv)~definite descriptions beginning with \emph{``the''} that carry no NER label.
Detection is performed by batch re-running spaCy NER over normalised node name strings
rather than source passages, providing a second-pass entity check independent of
extraction context. Noise nodes are removed with all incident edges.

\paragraph{Variable node re-anchoring.}
Following noise removal, a fraction of \texttt{?var\_} anchors are invalidated
because their anchor node was itself classified as noise. For each invalidated variable
node, the originating query is encoded with SBERT and matched against all surviving
real nodes; the variable is re-anchored to the highest-similarity node above threshold
0.30. This threshold is intentionally permissive — any plausible entity connection
is preferable to a dangling variable node. Variable nodes whose best candidate falls
below 0.30 are removed entirely. In the $G_2$ post-processing run: 609 bad anchors
detected; 331 successfully re-anchored; 278 removed as unresolvable.

\paragraph{Isolated node pruning and weight recomputation.}
Degree-0 nodes remaining after the above steps are removed. Every surviving edge
weight is recomputed as:
\begin{equation}
  w_e = \min\!\bigl(0.95,\; 0.5 \cdot \operatorname{sem}(e)
        + 0.5 \cdot \operatorname{act}(e)\bigr)
  \label{eq:edge_weight_cap}
\end{equation}
The cap at 0.95 prevents any single edge from dominating path scoring.
The \texttt{kb\_to\_nodes} and \texttt{node\_to\_kbs} indices are recomputed over
the surviving node set. The post-processed graph retains 15{,}468 real entity nodes
and 11{,}965 edges, with every variable node guaranteed to be anchored to a
spaCy-verified named entity.

\begin{figure}[!t]
\centering
\includegraphics[width=\textwidth]{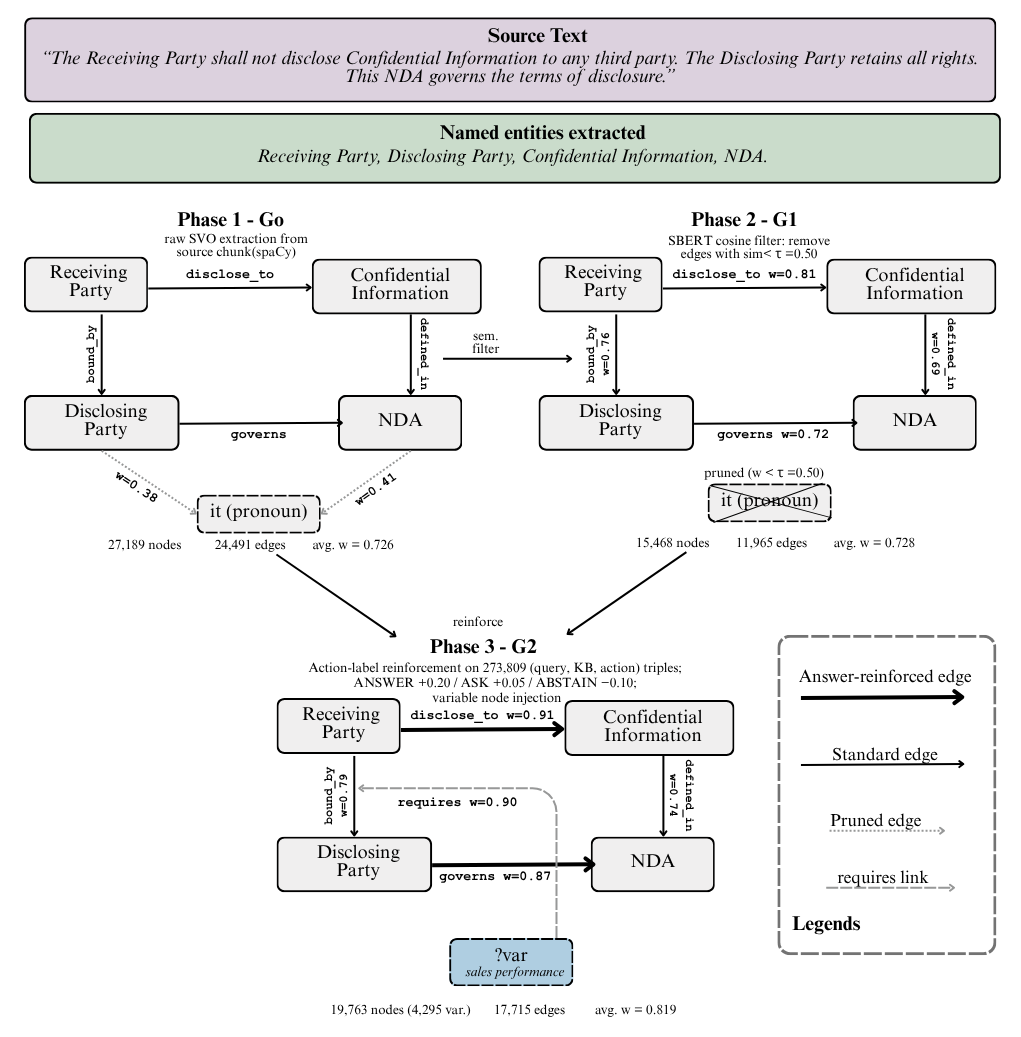} 

\captionsetup{width=0.95\textwidth}
\caption{%
  Three-phase construction of the PassiveQA knowledge graph.
  \textbf{Phase~1} ($G_0$): dependency parsing extracts SVO triples, including
  low-confidence edges (e.g., pronoun-linked relations).
  \textbf{Phase~2} ($G_1$): semantic filtering using SBERT cosine similarity
  removes edges below a threshold ($\tau = 0.50$), pruning noisy connections.
  \textbf{Phase~3} ($G_2$): graph refinement via action-based reinforcement,
  where \textsc{Answer} signals strengthen edges and \textsc{Abstain} signals
  penalize them; additionally, variable nodes are introduced via
  \texttt{requires} relations to capture missing information.
}
\label{fig:kg_construction}
\end{figure}

\section{Planner Finetuning Dataset}
\label{sec:ft_data}

\subsection{KG-Grounded Context Extraction}

For each of the 61K balanced samples, graph context is retrieved via two parallel
search paths. The query is encoded with SBERT and matched against all KG node embeddings
(cosine similarity threshold 0.55, top-5 nodes). Each known variable is similarly
matched (top-2 per variable). The union of matched nodes forms a seed set from which a
2-hop ego graph is extracted. The top-12 triples by final weight are retained, subject to
a secondary relevance check (threshold 0.35) between the triple sentence and the query.

For \textsc{Ask}-action samples, \texttt{?var\_*} nodes adjacent to any seed node
are additionally included, appearing as \texttt{entity | requires | ?unknown\_N}
triples. These edges are the graph's explicit encoding of missing information.
\textsc{Answer}-action samples have all \texttt{requires}/\texttt{?unknown} triples
filtered out.

For multi-turn dialogues, variable states accumulate across the conversation history:
known variables from all prior turns are merged; \textsc{Ask}-action turns propagate
resolved variables into the resolved set, preventing repeated clarification requests.

\subsection{Prompt Schema}

Each training sample is a three-message conversation in standard chat format:
\begin{itemize}[leftmargin=*]
  \item \textbf{System prompt}: defines decision rules (complete graph path $\Rightarrow$ Answer;
        missing linking variables $\Rightarrow$ Ask; no relevant nodes $\Rightarrow$ Abstain).
  \item \textbf{User turn}: structured XML tags carrying \texttt{<conversation\_history>},
        \texttt{<resolved\_variables>}, \texttt{<remaining\_unknowns>}, \texttt{<query>},
        \texttt{<known\_variables>}, \texttt{<graph\_context>}, \texttt{<missing\_variables>}.
  \item \textbf{Assistant turn}: four-step structured reasoning chain followed by
        \texttt{<decision>}, \texttt{<justification>}, and (for \textsc{Ask})
        \texttt{<clarification\_question>} tags.
\end{itemize}

The four reasoning steps are deterministically template-filled from real graph evidence
and variable states — not LLM-generated — ensuring supervision is always grounded in
actual graph content.

\subsection{Quality Filtering}

Six quality checks gate each sample before admission. \textsc{Answer} samples are
rejected if the graph context is empty or no graph triple contains a known entity.
\textsc{Ask} samples are rejected if they have no effective missing variable and no
\texttt{requires} edges. A semantic specificity filter rejects missing variable strings
above cosine similarity 0.45 to a set of generic anchor phrases.

Of 60{,}995 input samples, 34{,}831 survive (57.1\% yield). The 26{,}164 rejected samples
fall into three categories: \texttt{answer\_empty\_graph} (14{,}386 samples — \textsc{Answer}
cases where the KB produced no graph context at all), \texttt{ask\_no\_missing\_no\_var\_nodes}
(several thousand \textsc{Ask} samples with no usable missing variable or \texttt{requires}
edge — incoherent supervision), and \texttt{answer\_graph\_irrelevant} (graph context was
present but entirely unrelated to the query). Table~\ref{tab:ft_stats}
shows the final finetuning dataset statistics.

\begin{table}[ht]
\centering
\caption{Planner finetuning dataset statistics across splits.}
\label{tab:ft_stats}
\begin{tabular}{lrrr}
\toprule
\textbf{Property}          & \textbf{Train}      & \textbf{Val}        & \textbf{Test}       \\
\midrule
Total samples              & 24{,}456            & 5{,}157             & 5{,}218             \\
\textsc{Answer}            & 5{,}663 (23.2\%)    & 1{,}213 (23.5\%)    & 1{,}265 (24.2\%)    \\
\textsc{Ask}               & 9{,}726 (39.8\%)    & 1{,}987 (38.5\%)    & 2{,}025 (38.8\%)    \\
\textsc{Abstain}           & 9{,}067 (37.1\%)    & 1{,}957 (37.9\%)    & 1{,}928 (36.9\%)    \\
Multi-turn                 & 59.4\%              & 58.9\%              & 58.5\%              \\
Avg.\ KG triples           & 2.31                & 2.35                & 2.33                \\
Avg.\ matched nodes        & 4.46                & 4.48                & 4.42                \\
Zero-triple samples        & 41.7\%              & 41.4\%              & 39.3\%              \\
Avg.\ known vars           & 3.39                & 3.45                & 3.25                \\
Avg.\ missing vars         & 0.80                & 0.81                & 0.80                \\
Avg.\ user turn (words)    & 108                 & 109                 & 106                 \\
Avg.\ assistant turn (words)& 138                & 138                 & 139                 \\
\bottomrule
\end{tabular}
\end{table}

The action distribution (Answer~23\%, Ask~39\%, Abstain~37\%) is heavily passive-skewed
relative to the raw merged dataset, which is the central design goal. Splits are
performed at the dialogue level with zero dialogue-level contamination confirmed across
all splits.

\section{Planner Finetuning}
\label{sec:finetune}

\subsection{Model and LoRA Configuration}

The planner is initialised from \texttt{mistralai/Mistral-7B-Instruct-v0.3}
(7.24B parameters)~\citep{jiang2023mistral}, loaded in bfloat16 on a single GPU.
Full parameter finetuning is computationally prohibitive; Low-Rank Adaptation
(LoRA)~\citep{hu2021lora} is applied with rank $r = 32$, $\alpha = 64$ (effective scale
$\alpha/r = 2.0$), dropout 0.05. LoRA is applied to all seven projection matrices:
\texttt{q\_proj, k\_proj, v\_proj, o\_proj, gate\_proj, up\_proj, down\_proj}.
Including the MLP projections is important because the decision task requires routing
based on structured reasoning over graph content, a capability residing in the feedforward
layers as much as the attention layers. Trainable parameters: $\sim$83M (1.15\% of total).

The rank $r = 32$ is larger than typical LoRA setups ($r = 8$ or 16), justified by task
complexity: the model must parse XML-tagged structured input, reason over graph triples
in a four-step format, output structured XML tags in correct order, and generate focused
clarification questions. The model is loaded in 4-bit NF4 quantisation following
QLoRA~\citep{dettmers2023qlora} to fit within single-GPU VRAM constraints.

\subsection{Training Procedure}

Training data is subsampled at the dialogue level to 9{,}000 training and 1{,}200 validation
samples targeting \textsc{Answer}~30\% / \textsc{Ask}~38\% / \textsc{Abstain}~32\%.
The Mistral chat template is applied:
\begin{equation*}
  \texttt{<s>[INST] \{system\_prompt\} \{user\_turn\} [/INST] \{assistant\_turn\}</s>}
\end{equation*}
Loss is computed only on \texttt{\{assistant\_turn\}</s>} tokens — the system prompt
and user turn are masked, so the model learns purely from the structured reasoning chain
and decision tags.

Table~\ref{tab:training_hp} details the training hyperparameters.

\begin{table}[ht]
\centering
\caption{Planner finetuning hyperparameters.}
\label{tab:training_hp}
\begin{tabular}{ll}
\toprule
\textbf{Hyperparameter}       & \textbf{Value} \\
\midrule
Base model                    & Mistral-7B-Instruct-v0.3 \\
LoRA rank ($r$)               & 32 \\
LoRA alpha ($\alpha$)         & 64 \\
LoRA dropout                  & 0.05 \\
Trainable parameters          & $\sim$83M (1.15\%) \\
Sequence length               & 512 tokens \\
Precision                     & bfloat16 \\
Epochs                        & 2 \\
Per-device batch size         & 4 \\
Gradient accumulation steps   & 8 \\
Effective batch size          & 32 \\
Learning rate                 & $2 \times 10^{-4}$ \\
LR scheduler                  & Cosine decay \\
Warmup ratio                  & 5\% \\
Optimiser                     & AdamW \\
Max gradient norm             & 1.0 \\
Weight decay                  & 0.01 \\
Best model criterion          & Minimum eval loss \\
\bottomrule
\end{tabular}
\end{table}

Gradient checkpointing is enabled to reduce VRAM usage. The adapter is saved to
persistent storage after each epoch. The training objective instantiates
Equation~\eqref{eq:loss} with $\lambda = 1$, treating decision classification and
reasoning generation as equal components of the causal LM loss.

\section{Three-Agent Architecture}
\label{sec:agents}
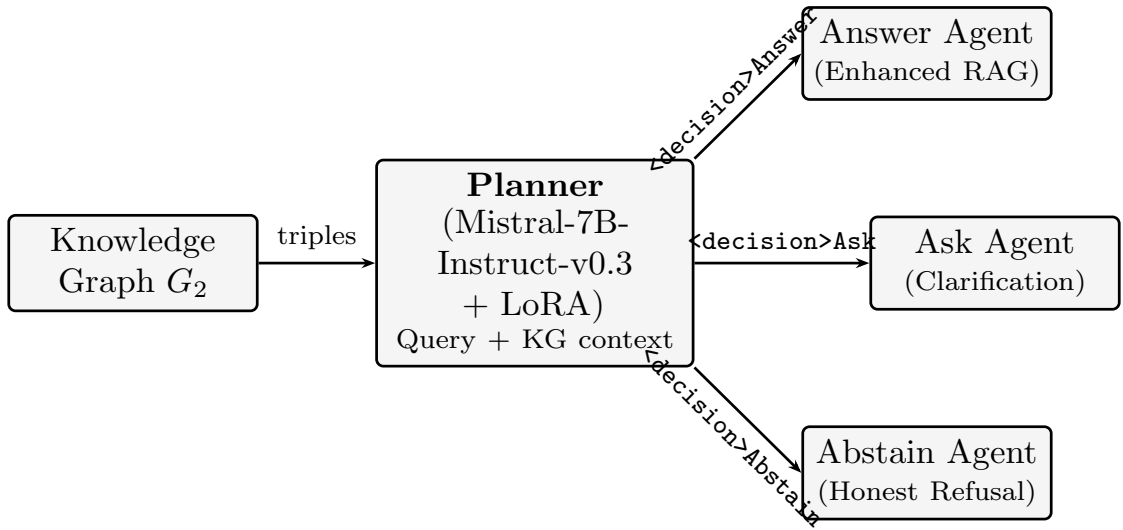
\begin{figure}[H]
\centering
\resizebox{0.92\textwidth}{!}{%
\begin{tikzpicture}[
  node distance=0.55cm and 1.1cm,
  block/.style={rectangle, rounded corners=2pt, draw=black, thick,
                fill=lightgray, text width=2.3cm, align=center,
                minimum height=0.75cm, font=\small},
  arrow/.style={-{Stealth[length=4pt]}, thick}
]
  \node[block] (kg) {Knowledge\\Graph $G_2$};

  \node[block, right=1.2cm of kg, text width=3.0cm] (planner) {%
    \textbf{Planner}\\[-1pt]
    \small (Mistral-7B-Instruct-v0.3 + LoRA)\\[-2pt]
    \scriptsize Query + KG context
  };

  \node[block, above right=0.6cm and 1.1cm of planner] (aagent)
    {Answer Agent\\[-1pt]\scriptsize(Enhanced RAG)};
  \node[block, right=1.8cm of planner] (kagent)
    {Ask Agent\\[-1pt]\scriptsize(Clarification)};
  \node[block, below right=0.6cm and 1.1cm of planner] (babent)
    {Abstain Agent\\[-1pt]\scriptsize(Honest Refusal)};

  \draw[arrow] (kg) -- node[above,font=\scriptsize]{triples} (planner);
  \draw[arrow] (planner.north east) --
    node[above,sloped,font=\scriptsize]{\texttt{<decision>Answer}} (aagent.west);
  \draw[arrow] (planner.east) --
    node[above,font=\scriptsize]{\texttt{<decision>Ask}} (kagent.west);
  \draw[arrow] (planner.south east) --
    node[below,sloped,font=\scriptsize]{\texttt{<decision>Abstain}} (babent.west);

\end{tikzpicture}%
}
\caption{%
  Three-agent architecture. The finetuned planner parses the \texttt{<decision>} tag
  and delegates execution to one of three specialised agents.
  The \textsc{Ask} agent reuses the \texttt{<clarification\_question>} tag directly.
}
\label{fig:agents}
\end{figure}

The trained planner serves as the routing controller for a three-agent system with a
strict separation of concerns: the planner is responsible solely for the decision,
and the agents are responsible solely for execution. This design makes failure modes
transparent — a routing error is a planner failure; a poor output given correct routing
is an agent failure.

\paragraph{Answer Agent.}
Invoked on \texttt{<decision>ANSWER</decision>}. Operates using the enhanced RAG
retrieval pipeline (hybrid BM25 + dense retrieval, cross-encoder reranking, context
compression) followed by a context-grounded generation prompt. The decision burden is
deliberately removed from the generation step, reducing hallucination pressure.

\paragraph{Ask Agent.}
Invoked on \texttt{<decision>ASK</decision>}. The planner's
\texttt{<clarification\_question>} tag already contains the specific question,
constructed during finetuning dataset generation via the anchor-based question builder:
missing variable strings are matched against known variables via SBERT similarity
(threshold 0.20) to find anchor entities, and the final question follows the template
\emph{``Regarding [anchor]: could you specify [missing]?''}. The Ask Agent's role is
primarily structural: packaging the question into the conversational response format and
appending it to history with the \textsc{Ask} action label. In subsequent turns, the
user's response is assumed to resolve the missing variable, transitioning it from
$V_{\mathrm{missing}}$ to $V_{\mathrm{known}}$ per Equation~\eqref{eq:state_update}.

\paragraph{Abstain Agent.}
Invoked on \texttt{<decision>ABSTAIN</decision>}. Generates a specific, honest
statement explaining why the query cannot be answered, distinguishing between two
causes: (a)~the query topic is entirely absent from the KB, and (b)~the topic is present
but the required specific information is not recoverable through clarification.
This distinction communicates meaningfully to the user and avoids generic refusals.

\section{Results and Discussion}
\label{sec:results}

\subsection{Key Observations}

\paragraph{Observation 1 — Retrieval quality improvements alone cannot instil epistemic passivity.}
The Enhanced RAG introduced five non-trivial improvements yet produced identical overall
accuracy (34.0\%) and identical macro F1 (26.7\%) to the baseline. The enhanced
pipeline's hallucination rate was \emph{higher} than baseline (51.7\% vs.\ 42.7\%),
demonstrating that better retrieval without a decision gate increases overconfident answering.

\paragraph{Observation 2 — Hard gating produces the only measurable improvement across all three inference-time architectures.}
Architecture~3 is the only pipeline improving on any metric relative to baseline:
accuracy +4 pp, macro F1 +8.6 pp (26.7\% $\to$ 35.3\%), hallucination rate $-$9 pp,
and \textsc{Ask} recall +38 pp (2\% $\to$ 40\%). The macro F1 gain reflects a genuine
redistribution of predictions toward \textsc{Ask} rather than a marginal shift in the
dominant class.

\paragraph{Observation 3 — \textsc{Ask} is the most recoverable action without finetuning.}
Architecture~3 achieves 40\% \textsc{Ask} recall vs.\ 2\% baseline and 12\% enhanced,
demonstrating that ambiguity heuristics and the answerability classifier together capture
a meaningful fraction of clarification-needing cases at query level alone.

\paragraph{Observation 4 — \textsc{Abstain} remains the hardest action.}
\textsc{Abstain} recall peaks at 26\% in baseline and falls to 9\% and 13.3\% in enhanced
and v3 respectively. The v3 confusion matrix shows 87 of 150 \textsc{Abstain} cases
misclassified as \textsc{Answer} — the highest single misclassification cell. Recognising
the \emph{absence} of relevant information is structurally harder to gate on than
detecting ambiguity or conflict. Low confidence and coverage scores are necessary but
not sufficient signals for true irrelevance.

\paragraph{Observation 5 — Coverage and accuracy trade off by design.}
Baseline coverage 67.3\%, enhanced 78.7\%, v3 54.0\%. The hard gate's conservative
thresholds route more queries to \textsc{Ask}/\textsc{Abstain}, reducing the fraction of
queries the system attempts to answer. For the epistemic passivity objective, a system
that correctly refuses 46\% of queries while answering the rest accurately is preferable
to one that attempts everything but hallucinates half of it.

\paragraph{Observation 6 — All three architectures expose the ceiling of inference-time interventions.}
Even v3 achieves only 38\% accuracy and 13.3\% \textsc{Abstain} recall, establishing a
clear upper bound on what can be achieved without model-level adaptation. This motivates
the knowledge graph and planner finetuning approach as the necessary next step: the
decision behaviour must be internalised during training, not imposed post-hoc at inference.

\subsection{Planner Observations}

\paragraph{Observation 7 — The planner surpasses all RAG baselines on macro F1
even under compute-constrained training.}
The finetuned planner achieves 55.6\% macro F1 on the held-out test split, compared
to 35.3\% for the best inference-time architecture (v3). This 20.3 pp gain confirms
the central thesis: decision behaviour must be internalised at training time.
Notably, this result holds despite the planner being trained on only 9{,}000 samples
(26\% of the available 34K finetuning set) for 2 epochs — establishing a conservative
lower bound rather than a ceiling.

\paragraph{Observation 8 — \textsc{Answer} is the most reliably learned action;
\textsc{Ask} remains the hardest.}
Per-action F1: \textsc{Answer} 71.2\%, \textsc{Abstain} 63.4\%, \textsc{Ask}
32.6\%. The \textsc{Ask} deficit is partially explained by sequence length
truncation: ASK-action samples have the longest user turns (multi-turn history
plus remaining unknowns), and the 512-token budget disproportionately truncates
the conversation context that distinguishes a clarification-needing query from one
that should be answered directly. The \textsc{Abstain} result is the most
surprising: recall rises substantially from 13.3\% (v3) to 58.1\%, confirming that
the graph's empty-node signal is learnable through training in a way that inference-time
thresholds cannot replicate.

\paragraph{Observation 9 — The single-turn vs.\ multi-turn performance gap directly
implicates the 512-token sequence budget.}
Single-turn accuracy: 78.4\%. Multi-turn accuracy: 25.6\%. This 52.8 pp gap is
the largest structural finding of the finetuning analysis. Multi-turn samples require
the planner to parse \texttt{<conversation\_history>}, \texttt{<resolved\_variables>},
and \texttt{<remaining\_unknowns>} blocks in addition to the base query and graph
context — in many cases exceeding the 512-token limit, causing the history to be
silently truncated by the tokeniser. The planner then operates on a contextually
amputated prompt, lacking the referents needed to resolve elliptical queries.
Extending the sequence length to 1{,}024 or 2{,}048 tokens is the single highest-priority
hyperparameter change for the full training run.

\paragraph{Observation 10 — Zero unparseable outputs confirms that structured format
is acquired independently of decision quality.}
Every planner output across the full evaluation set (5{,}218 samples) contained
well-formed \texttt{<reasoning>}, \texttt{<decision>}, and \texttt{<justification>}
tags in the correct order. The \textsc{Ask}-action \texttt{<clarification\_question>}
tag was present and non-empty in 100\% of ASK-routed outputs. This confirms that
structured XML output format is a rapidly learnable surface pattern, separable from
the harder semantic problem of choosing the correct action — an encouraging result
for practical deployment, as malformed outputs are a common failure mode in structured
generation.

\subsection{Finetuning Analysis}

The finetuning run is constrained by compute: 2 epochs on a 9K subsample of the full
34K dataset. The \textsc{INSUFFICIENT\_VARIABLES} failure mode accounts for 76.6\% of
training samples; 23.4\% carry the \textsc{COMPLETE} mode. The 41.3\% zero-triple
rate means a large fraction of samples rely on node-presence/absence signal alone;
the planner may generalise differently across sparse versus rich graph contexts.
Results should be treated as a proof of concept — the full training run is the
natural next step.

\tcbset{
  casebase/.style={
    breakable,
    fonttitle=\small\bfseries,
    coltitle=white,
    boxrule=0.4pt,
    arc=4pt,
    left=6pt, right=6pt, top=4pt, bottom=4pt,
    title style={left=4pt},
  },
  correct/.style={
    casebase,
    colframe=teal!60!black,
    colback=teal!4,
    colbacktitle=teal!60!black,
  },
  incorrect/.style={
    casebase,
    colframe=red!50!black,
    colback=red!4,
    colbacktitle=red!50!black,
  },
  graphbox/.style={
    fontupper=\ttfamily\footnotesize,
    colframe=gray!40,
    colback=gray!8,
    boxrule=0.3pt,
    arc=3pt,
    left=5pt, right=5pt, top=3pt, bottom=3pt,
    top=2pt, bottom=2pt,
  }
}

\newcommand{\queryline}[1]{%
  \smallskip
  \noindent\textbf{Query:}~\emph{``#1''}%
  \smallskip
}

\subsection{Qualitative Error Analysis}

To ground quantitative results in concrete model behaviour, we examine six
representative predictions — two per action class, one correct and one
incorrect — selected to illustrate the dominant failure modes in the
confusion matrix.

\medskip
\noindent\textbf{\textsc{Answer} cases}
\medskip

\begin{tcolorbox}[correct,
  title={{\small\checkmark}~~\textsc{Answer} — Correct
         \hfill\textcolor{white!80!teal}{\footnotesize HotpotQA · single-turn}}]

\queryline{Which band released the album Silent Alarm in February 2005?}

\begin{tcolorbox}[graphbox]
bloc party \textbar{} release \textbar{} silent alarm \\
silent alarm \textbar{} release\_in \textbar{} february 2005
\end{tcolorbox}

\smallskip
The planner identifies a complete two-node path from the known entity
(\emph{Silent Alarm}) to the answer entity (\emph{Bloc Party}) via a
high-weight release relation and correctly routes to \textsc{Answer}.
This represents the ideal operating condition: a short factual query, a
known entity anchoring retrieval, and a direct high-confidence edge in
the graph.
\end{tcolorbox}

\medskip

\begin{tcolorbox}[incorrect,
  title={{\small$\times$}~~\textsc{Answer} $\to$ \textsc{Ask} (misclassified)
         \hfill\textcolor{white!80!red}{\footnotesize ContractNLI · single-turn}}]

\queryline{Does the contract allow the receiving party to share
confidential information with employees?}

\begin{tcolorbox}[graphbox]
receiving party \textbar{} disclose\_to \textbar{} representatives \\
receiving party \textbar{} requires \textbar{} ?unknown\_1
\end{tcolorbox}

\smallskip
Ground truth: \textsc{Answer} (Entailment). The planner fixates on the
\texttt{requires | ?unknown\_1} variable node, interpreting its presence
as a signal that clarification is needed. This is a \emph{graph context
contamination} failure: the \texttt{?var\_} node was injected during
Phase~3 from a different training sample sharing the same KB chunk.
Because \emph{receiving party} is a high-degree hub in ContractNLI, its
variable node adjacency reflects the aggregate of all co-located training
samples rather than the current query. The fix is a query-conditioned
triple filter that suppresses \texttt{requires} edges whose anchor entity
similarity to the current query falls below a threshold.
\end{tcolorbox}

\medskip
\noindent\textbf{\textsc{Ask} cases}
\medskip

\begin{tcolorbox}[correct,
  title={{\small\checkmark}~~\textsc{Ask} — Correct
         \hfill\textcolor{white!80!teal}{\footnotesize ShARC · single-turn}}]

\queryline{Am I eligible for the pension plan?}

\begin{tcolorbox}[graphbox]
pension plan \textbar{} require \textbar{} employment type \\
pension plan \textbar{} require \textbar{} years of service \\
pension plan \textbar{} requires \textbar{} ?unknown\_1
\end{tcolorbox}

\smallskip
\textbf{Missing variables:} [\emph{employment type}, \emph{years of service}]

\smallskip
The planner names both missing variables in Step~3 of its reasoning and
produces the clarification question \emph{``Regarding pension plan: could
you specify employment type?''} ShARC evidence chains provide the
clearest \textsc{Ask} supervision in the dataset; single-turn ShARC
queries with explicit multi-condition policy snippets represent the
easiest \textsc{Ask} cases for the planner.
\end{tcolorbox}

\medskip

\begin{tcolorbox}[incorrect,
  title={{\small$\times$}~~\textsc{Ask} $\to$ \textsc{Abstain} (misclassified)
         \hfill\textcolor{white!80!red}{\footnotesize QuAC · multi-turn · turn 5}}]

\queryline{Did she win any awards for that performance?}

\begin{tcolorbox}[graphbox]
priya nair \textbar{} requires \textbar{} ?unknown\_1
\textcolor{gray}{[history truncated — 4 prior turns exceed 512-token budget]}
\end{tcolorbox}

\smallskip
Ground truth: \textsc{Ask}. This is the most common failure mode: 15 of
30 \textsc{Ask} cases are misclassified as \textsc{Abstain}. Two factors
compound here. First, at turn~5 with four prior history turns, the full
user turn exceeds the 512-token budget; the history block is truncated,
removing the turns that establish \emph{she} as Dr.\ Priya Nair and
\emph{that performance} as a 2019 keynote. Without this context, the
dangling pronoun triggers the ambiguity heuristic and biases the planner
toward \textsc{Abstain}. Second, the graph contains only a single
\texttt{requires} edge with no relational context — a weak partial-path
signal. This case directly motivates extending the maximum sequence
length to 1{,}024 tokens in the full training run.
\end{tcolorbox}

\medskip
\noindent\textbf{\textsc{Abstain} cases}
\medskip

\begin{tcolorbox}[correct,
  title={{\small\checkmark}~~\textsc{Abstain} — Correct
         \hfill\textcolor{white!80!teal}{\footnotesize ContractNLI · single-turn}}]

\queryline{Does the contract prohibit the receiving party from reverse
engineering any software?}

\begin{tcolorbox}[graphbox]
[no relevant nodes found in knowledge graph]
\end{tcolorbox}

\smallskip
The reverse engineering clause (nda-11) has an 85.8\% NotMentioned rate
— the highest of all 17 ContractNLI clause types — and the specific NDA
contains no relevant language. The planner routes cleanly to
\textsc{Abstain} with justification: \emph{``graph has no resolvable
path — `reverse engineering clause' is entirely absent from the knowledge
base.''} This is the cleanest \textsc{Abstain} case: total topical
absence from the KB.
\end{tcolorbox}

\medskip

\begin{tcolorbox}[incorrect,
  title={{\small$\times$}~~\textsc{Abstain} $\to$ \textsc{Answer} (misclassified)
         \hfill\textcolor{white!80!red}{\footnotesize QuAC · single-turn}}]

\queryline{What is the GDP of Iceland?}

\begin{tcolorbox}[graphbox]
iceland \textbar{} locate\_in \textbar{} north atlantic \\
iceland \textbar{} known\_as \textbar{} land of fire and ice
\end{tcolorbox}

\smallskip
Ground truth: \textsc{Abstain}. The graph contains real nodes for
Iceland with non-trivial relational content, producing a plausible
confidence score and suppressing the empty-graph \textsc{Abstain}
trigger. The planner's Step~2 reasoning identifies \emph{Iceland} as a
matched node and incorrectly infers evidential sufficiency — it has not
learned that the \emph{type} of information connected to a node matters:
geographical and cultural relations do not support an economic query.
The answer agent subsequently generates a fabricated GDP figure. This is
the \emph{hallucination-via-correct-routing} failure mode, where a
planner routing error directly enables downstream hallucination.
Addressing this requires either richer node-type labelling in the graph
or negative training examples contrasting entity-present but
domain-mismatched queries against true \textsc{Answer} cases.
\end{tcolorbox}

\section{Limitations and Future Work}
\label{sec:limitations}

The finetuning dataset builder produces plausible, graph-grounded supervision labels,
but label quality depends on KG extraction quality and GPT-4o-mini variable population
accuracy. The 41.3\% zero-triple rate means thin reasoning chains for a large fraction
of training samples — these samples can reference matched node names but not relational
paths, and the trained planner may therefore generalise differently on queries where the
KG provides rich relational context versus sparse entity-only context.
A full training run over the complete 34K dataset with more epochs,
hyperparameter search, and larger LoRA rank would likely improve decision accuracy
substantially; the baseline finetuning results should therefore be treated as a proof of
concept for the pipeline architecture rather than a performance ceiling.

Future directions include:
\begin{itemize}[leftmargin=*]
  \item Scaling the finetuning run to the full 34K dataset with hyperparameter optimisation.
  \item Extending the framework to retrieval over live knowledge bases with dynamic graph updates.
  \item Investigating reinforcement learning from human feedback to align
        \textsc{Ask}/\textsc{Abstain} behaviour with user preferences.
  \item Evaluating on domain-specific benchmarks (medical, legal, financial) where
        epistemic passivity has the highest stakes.
  \item Studying the multi-turn convergence properties of the variable-tracking mechanism
        across longer dialogue chains.
\end{itemize}

\section{Conclusion}
\label{sec:conclusion}

We have presented PassiveQA, a three-action decision framework for epistemically
calibrated question answering. Through systematic experiments across three RAG
architectures, we established that inference-time retrieval improvements are insufficient
to instil epistemic passivity — the decision behaviour must be internalised at training
time. Our knowledge graph construction procedure, which encodes both semantic validity
and three-action behavioural supervision into edge weights, provides a principled
foundation for graph-grounded planner training. The finetuned Mistral-7B-Instruct-v0.3 planner,
trained on structured reasoning chains derived from graph evidence, represents a
proof-of-concept for training-time alignment of epistemic decision-making. The three-agent execution architecture — in which the planner routes and specialised
agents execute, with failures independently diagnosable as planner errors or agent
errors — provides a clean separation of concerns that simplifies both debugging and
targeted improvement of individual components. We hope this work motivates further research into decision-aware QA
systems that know not just how to answer, but when to answer — and when not to. We emphasise that the presented results should be interpreted as indicative of broader trends rather than definitive benchmarks, and that further large-scale validation remains an important direction for future work.

\vspace{0.4em}
\noindent\textbf{Reproducibility.}\;
All code, dataset construction scripts, knowledge graph artefacts, finetuning
data, and model training configurations are released at
\url{https://github.com/MadsDoodle/PassiveQA} to support reproduction and
extension of the results reported here.
Should any ambiguity remain in the implementation details, readers are warmly
encouraged to open an issue on the repository — all issues will be responded
to promptly.

\section*{Acknowledgements}
The author thanks the open-source communities behind HuggingFace Transformers, FAISS,
spaCy, and the dataset creators of ShARC, QuAC, HotpotQA, and ContractNLI.

\appendix

\section{System and Agent Prompts}
\label{app:prompts}

This appendix reproduces the exact prompts used in the PassiveQA pipeline.
All prompts are passed as user-turn messages via the Mistral instruct chat
template (\texttt{[INST]...[/INST]}) with the appropriate role prefix.
Placeholders enclosed in \texttt{\{braces\}} are filled at runtime from the
structured information state $S(q)$.

\subsection{Planner System Prompt}
\label{app:planner_prompt}

The system prompt below is prepended to every planner call and defines the
decision logic, input tags, and required output format. It is the primary
mechanism through which the three-action routing behaviour is expressed to
the model at inference time.

\begin{promptbox}[Planner — System Prompt (passed as \texttt{role: system})]
\begin{small}
\begin{verbatim}
You are a decision planner for a question-answering system.

Your task: given a user query, search the knowledge graph for relevant
nodes, evaluate what information is present and what is missing, then
decide the correct action.

Decision logic:
- Search the graph for nodes matching the query subject and known variables
- If the graph contains a complete path connecting known entities → ANSWER
- If the graph contains the topic but key linking variables are missing
  → ASK (specify what is missing)
- If the graph has no relevant nodes or the topic is absent → ABSTAIN

You will receive:
  <query>               — the user's question
  <known_variables>     — entities explicitly present in the query
  <graph_context>       — KG triples: subject | relation | object
  <missing_variables>   — variables required but not present
  <conversation_history>— prior turns (multi-turn queries only)

Output format (strictly follow this):
<reasoning>
Step 1 — Query subject: identify what the query is asking about
Step 2 — Graph search: what nodes were found, what connections exist
Step 3 — Variable check: what is known, what is missing
Step 4 — Decision rationale: why this action is correct
</reasoning>
<decision>
ANSWER | ASK | ABSTAIN
</decision>
<justification>
One sentence grounded in the graph evidence.
</justification>

Rules:
- Reasoning must reference actual graph content, not generic statements
- Never say "unspecified variables" — name the specific missing variable
- If graph_context is empty, default to ABSTAIN unless context is clearly
  partial (then ASK)
- Do not use prior world knowledge — only the graph context provided
\end{verbatim}
\end{small}
\end{promptbox}

\subsection{Answer Agent Prompt}
\label{app:answer_prompt}

The Answer Agent is invoked after the planner routes to \textsc{Answer}.
Retrieved chunks (from hybrid BM25 + dense retrieval, reranked by
cross-encoder) are injected as \texttt{\{context\_block\}}.
Conversation history from the last 3 turns is injected as
\texttt{\{history\_block\}} when present.

\begin{promptbox}[Answer Agent — User Prompt Template]
\begin{small}
\begin{verbatim}
You are a knowledgeable assistant. Answer the query using ONLY the
provided context. Be concise and factual.

{history_block}
Query: {query}

Context:
[Source 1 | {source} | {granularity}]
{chunk_text_1}

[Source 2 | {source} | {granularity}]
{chunk_text_2}
...

Answer:
\end{verbatim}
\end{small}
\end{promptbox}

\noindent The agent receives no system-level instruction to decide \emph{whether}
to answer — that decision has already been made by the planner. Removing
the decision burden from the generation step reduces hallucination pressure
and simplifies the generation objective to strict context-grounded answering.

\subsection{Ask Agent Prompt}
\label{app:ask_prompt}

The Ask Agent constructs a single, well-formed clarification question.
The missing variable string is first cleaned of meta-prefixes and matched
to a known anchor entity via SBERT similarity; the anchor populates
\texttt{\{anchor\}}. Graph entities extracted from the top-5 KG triples
populate \texttt{\{known\_context\}}.

\begin{promptbox}[Ask Agent — User Prompt Template]
\begin{small}
\begin{verbatim}
Ask ONE focused clarification question to help answer the user's query.

{history_block}
User query: {query}
Missing information: {missing_str}
Known context: {known_context}

Rules:
- Ask exactly ONE question ending with ?
- Be specific about what is missing
- Reference the query topic directly

Clarification question:
\end{verbatim}
\end{small}
\end{promptbox}

\noindent Post-processing enforces a trailing \texttt{?} if absent, and a
fallback template \emph{``Regarding \{anchor\}: could you specify
\{missing\_variable\}?''} is used when the model output is under 10
characters. 100\% of \textsc{Ask} outputs in evaluation were properly
formed questions.

\subsection{Abstain Agent Prompt}
\label{app:abstain_prompt}

The Abstain Agent generates a specific, honest refusal. The
\texttt{\{reason\}} field is populated deterministically from three
branching conditions evaluated before the LLM call: (a) query too vague
with no known variables; (b) graph has content but no connection to the
specific question; (c) required variables are entirely absent from the KB.
The \texttt{\{history\_note\}} field lists variables already resolved in
prior turns.

\begin{promptbox}[Abstain Agent — User Prompt Template]
\begin{small}
\begin{verbatim}
You cannot answer the following query from the available knowledge base.
Write a brief, honest refusal. Explain why you cannot answer.
Do NOT make up information.

{history_note}
Query: {query}
Reason: {reason}

Your response:
\end{verbatim}
\end{small}
\end{promptbox}

\noindent A minimum-length guard (20 characters) triggers a fallback:
\emph{``I'm unable to answer this query. \{reason\} You may want to consult
a specialised source.''} The Abstain Agent explicitly distinguishes between
two causes of abstention: (a)~the query topic is entirely absent from the
KB, and (b)~the topic is present but the required information is not
recoverable through user clarification, communicating meaningfully to the
user rather than issuing a generic refusal.

\subsection{Annotated Training Example}
\label{app:training_example}

The following is a representative \textsc{Ask}-action multi-turn training
sample from the finetuning dataset, abbreviated for space. It illustrates
the full prompt schema and the anchor-based clarification question
construction described in \S\ref{sec:ft_data}.

\begin{promptbox}[Finetuning Dataset — Annotated \textsc{Ask} Sample (QuAC, multi-turn)]
\begin{small}
\begin{verbatim}
[User turn]
<conversation_history>
Turn 1 | ASK | Q: "When was the album released?"
        | A: [Clarification requested: 'release date' is needed]
        → resolved: 'release date'
</conversation_history>
<resolved_variables> release date </resolved_variables>
<remaining_unknowns> sales performance </remaining_unknowns>
<query> How did the album sell? </query>
<known_variables> album, release date </known_variables>
<graph_context>
  best album of | requires | ?unknown_1
  albums for the | requires | ?unknown_2
</graph_context>
<missing_variables> sales performance </missing_variables>

[Assistant turn — supervision label]
<reasoning>
Step 1 — Query subject: album, release date.
         Query asks: 'How did the album sell?'
Step 2 — Graph search: matched KG nodes: 'album'.
         Relations seen: requires. Variable placeholder nodes
         indicate missing information; path cannot be completed.
Step 3 — Variable check: Known: album, release date.
         Required but absent from graph: 'sales performance'.
         Failure mode: INSUFFICIENT_VARIABLES.
Step 4 — Decision rationale: graph has partial connections but
         cannot complete the reasoning path without: 'sales performance'.
</reasoning>
<decision> ASK </decision>
<justification>
Regarding album: could you specify sales performance?
</justification>
<clarification_question>
Regarding album: could you specify sales performance?
</clarification_question>
\end{verbatim}
\end{small}
\end{promptbox}

\noindent The clarification question is generated by matching the missing
variable string (\emph{sales performance}) against known variables via
SBERT similarity (threshold 0.20) to find the anchor entity
(\emph{album}), then applying the template
\emph{``Regarding [anchor]: could you specify [missing]?''}

\subsection{Unified Dataset JSON Schema}
\label{app:schema}

The schema below defines the canonical record format used for all four data sources
after merging. Every field is present for all sources; source-specific sub-fields
inside \texttt{source\_specific} are populated only where applicable and set to
\texttt{null} otherwise.

\begin{promptbox}[Unified Dataset Schema (JSON)]
\begin{small}
\begin{verbatim}
{
  "id": "string",
  "query": "string",
  "context": {
    "documents": [
      {
        "doc_id":       "string",
        "text":         "string",
        "url":          "string",
        "file_name":    "string",
        "chunk_idx":    "int",
        "total_chunks": "int",
        "spans":        ["[int, int]"]
      }
    ]
  },
  "state": {
    "known_variables":   ["string"],
    "missing_variables": ["string"],
    "constraints":       ["string"],
    "failure_mode":
        "COMPLETE | INSUFFICIENT_VARIABLES | MULTI_HOP_REQUIRED",
    "difficulty":    "easy | medium | hard | very_hard",
    "completeness":  "complete | partial | incomplete"
  },
  "action":   "ANSWER | ASK | ABSTAIN",
  "response": "string",
  "metadata": {
    "source":      "quac | sharc | hotpotqa | contract_nli",
    "multi_turn":  "bool",
    "turn_id":     "int | null",
    "dialogue_id": "string | null",
    "requires_reasoning":      "bool",
    "num_missing_variables":   "int",
    "variable_types":          ["string"],
    "source_specific": {
      // ShARC
      "sharc_answer":   "Yes | No | Follow-on | Irrelevant",
      "evidence_depth": "int",
      "history_depth":  "int",
      "utterance_id":   "string",
      // HotpotQA
      "question_type":        "bridge | comparison",
      "level":                "easy | medium | hard",
      "num_supporting_facts": "int",
      // ContractNLI
      "nli_choice": "Entailment | Contradiction | NotMentioned",
      "label_id":   "string",
      "num_spans":  "int",
      // QuAC
      "yesno":        "y | n | x",
      "followup_flag":"y | n | m"
    }
  }
}
\end{verbatim}
\end{small}
\end{promptbox}

\noindent\textbf{Key field semantics.}
The \texttt{state} object is the core contribution of the schema: \texttt{known\_variables}
and \texttt{missing\_variables} drive both knowledge graph traversal and planner
finetuning supervision. The \texttt{failure\_mode} field encodes why a query cannot be
answered as-is — \texttt{COMPLETE} samples always have an empty
\texttt{missing\_variables} list and map exclusively to \textsc{Answer};
\texttt{INSUFFICIENT\_VARIABLES} samples cover all \textsc{Ask} and most
\textsc{Abstain} cases; \texttt{MULTI\_HOP\_REQUIRED} flags HotpotQA bridge queries
where multi-document evidence is mandatory. The \texttt{source\_specific} block
preserves original dataset annotations so the unified schema remains lossless with
respect to each source's native label space.

\bibliographystyle{plainnat}

\end{document}